\documentclass[10pt,twocolumn,letterpaper]{article}

\usepackage{wacv}              

\usepackage[accsupp]{axessibility}  

\usepackage{amsmath}
\usepackage{enumitem}
\usepackage{mathtools}
\usepackage{amsthm}
\usepackage{multirow}
\usepackage{tikz}
\usepackage{pifont}
\usepackage{booktabs}
\usepackage[flushleft]{threeparttable}
\usepackage{algorithm}
\usepackage{algorithmic}

\definecolor{DeepBlue}{RGB}{0, 92, 171}
\definecolor{MediumBlue}{RGB}{100, 160, 200}
\definecolor{LightBlue}{RGB}{220, 238, 243}
\newcommand{\firsttone}[1]{\colorbox{DeepBlue!40}{#1}}
\newcommand{\secondtone}[1]{\colorbox{MediumBlue!40}{#1}}
\newcommand{\thirdtone}[1]{\colorbox{LightBlue}{#1}}

\newcommand{\cmark}{\ding{51}}
\newcommand{\xmark}{\ding{55}}

\definecolor{wacvblue}{rgb}{0.21,0.49,0.74}
\usepackage[pagebackref,breaklinks,colorlinks,allcolors=wacvblue]{hyperref}

\def\wacvPaperID{1634} 
\def\confName{WACV}
\def\confYear{2026}

\title{From Darkness to Detail: Frequency-Aware SSMs for Low-Light Vision}

\author{
    Eashan Adhikarla\\
    Lehigh University\\
    Bethlehem, Pennsylvania, USA\\
    {\tt\small eaa418@lehigh.edu}
\and
    Kai Zhang\\
    Lehigh University\\
    Bethlehem, Pennsylvania, USA\\
    {\tt\small kaz321@lehigh.edu}
\and
    Gong Chen\\
    Lenovo Research\\
    Chicago, Illinois, USA\\
    {\tt\small gochen24@lenovo.com}
\and
    John Nicholson\\
    Lenovo Research\\
    Raleigh, North Carolina, USA\\
    {\tt\small jnichol@lenovo.com}
\and
    Brian D. Davison\\
    Lehigh University\\
    Bethlehem, Pennsylvania, USA\\
    {\tt\small bdd3@lehigh.edu}
}

\begin{document}

\makeatletter
\let\@oldmaketitle\@maketitle
\renewcommand{\@maketitle}{\@oldmaketitle \vspace{-.75cm}
    \begin{center}
    \begin{minipage}{\textwidth}
        \centering
        \includegraphics[width=0.36\textwidth]{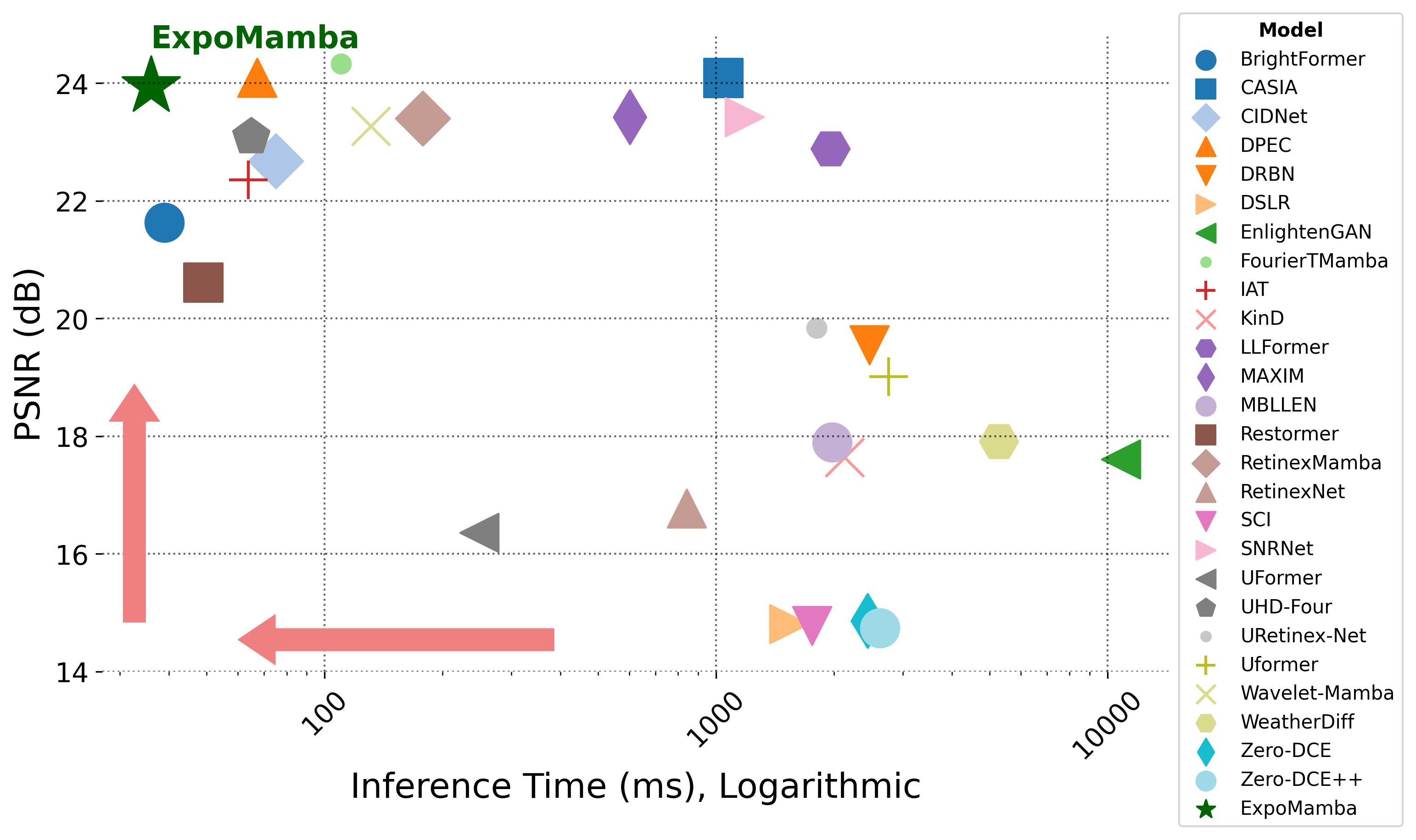}
        \includegraphics[width=0.36\textwidth]{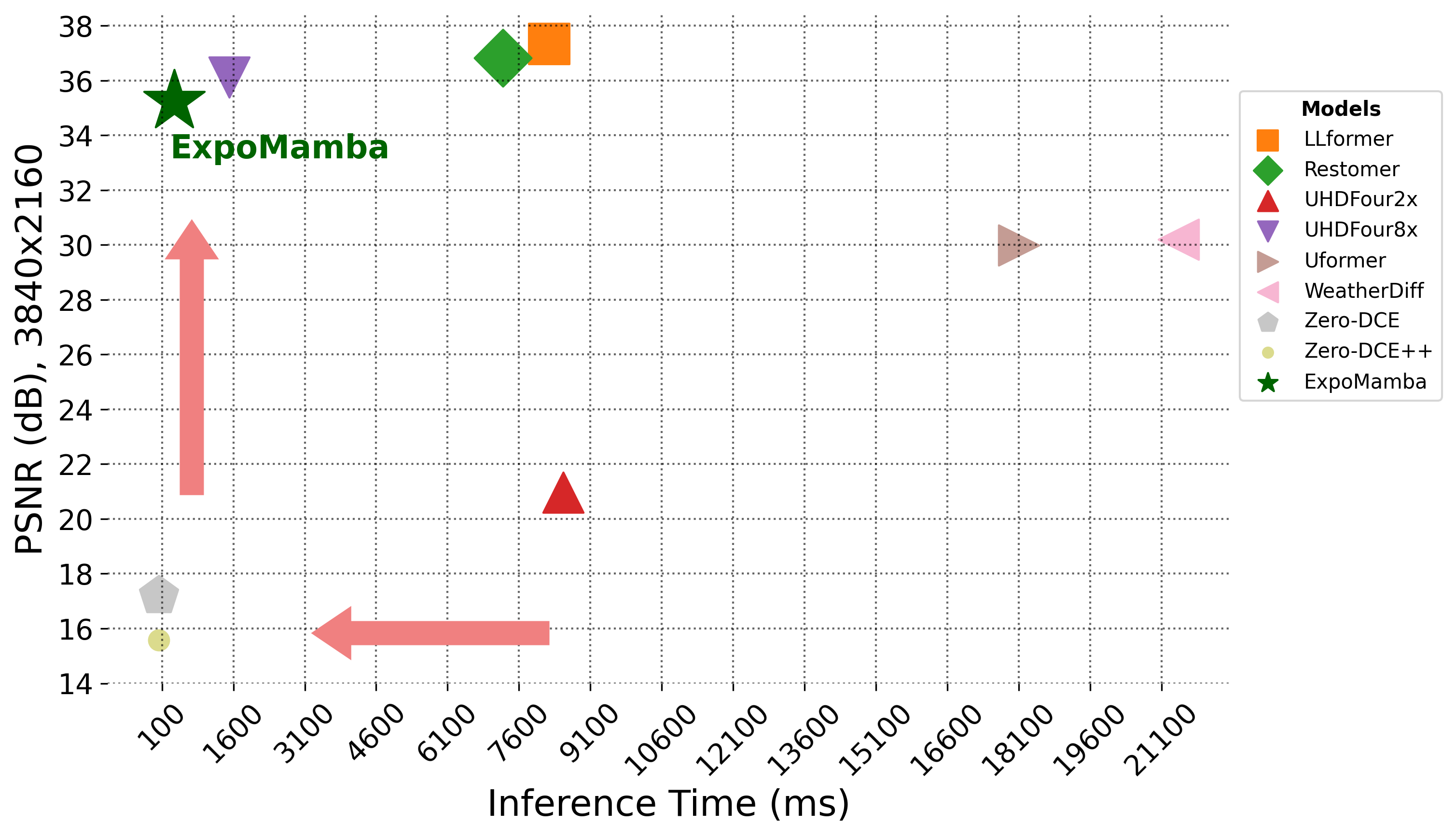}
        \includegraphics[width=0.25\textwidth]{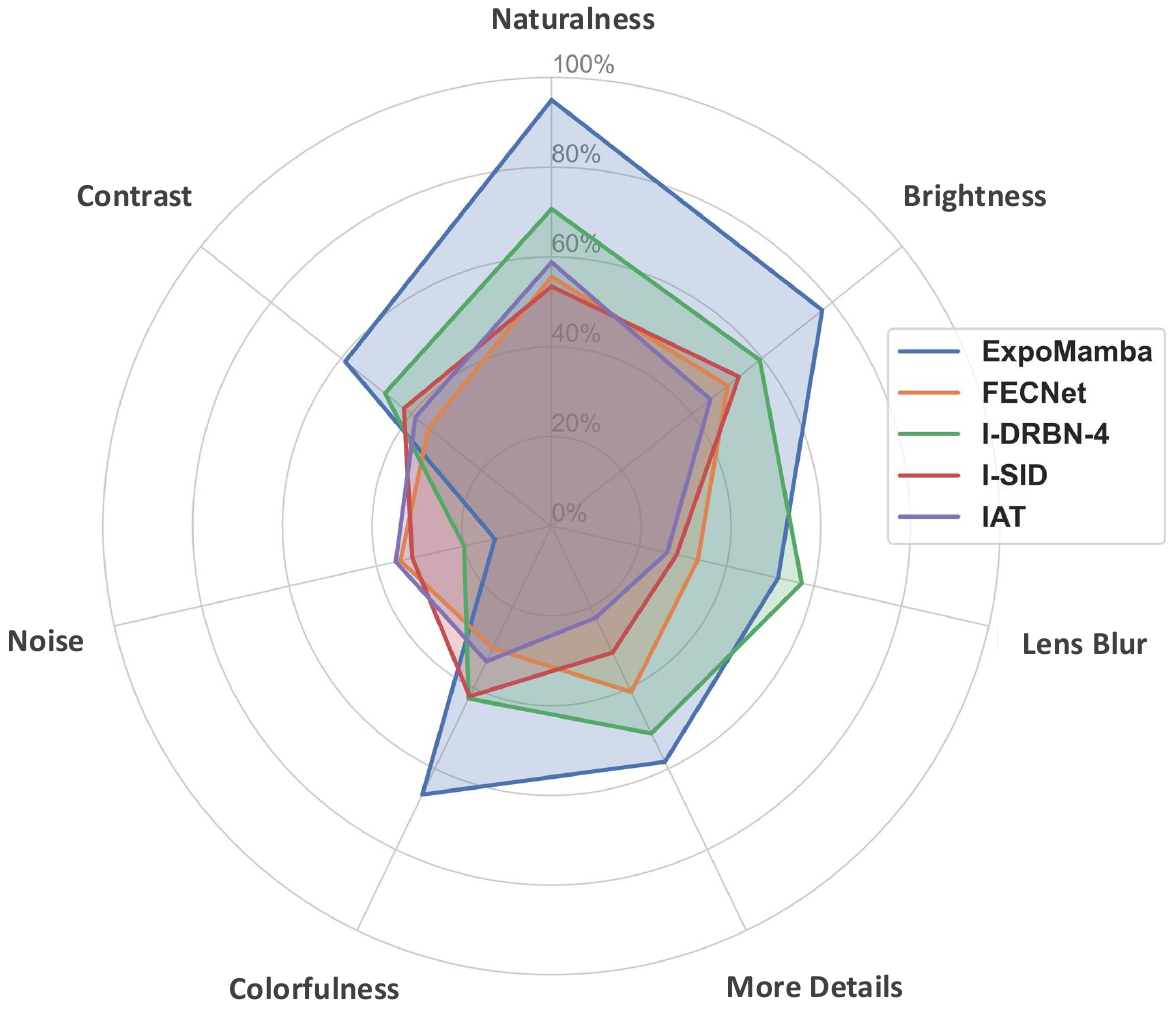}
        \captionof{figure}{\label{fig:teaser}[\textbf{Left:}~400x600; \textbf{Middle:}~3840x2160]~Logarithmic Scatter Plot of Inference Time vs.\ PSNR. Baselines that used ground-truth mean information to produce metrics were reproduced without such information for fairness, [\textbf{Right:}] Spider chart with comparison of ExpoMamba outputs with other SOTA light-weight models on perceptual realism.}
    \end{minipage}
    \end{center}
}
\makeatother




\maketitle
\begin{abstract}
    Low-light image enhancement remains a persistent challenge in computer vision, where state-of-the-art models are often hampered by hardware constraints and computational inefficiency, particularly at high resolutions. While foundational architectures like transformers and diffusion models have advanced the field, their computational complexity limits their deployment on edge devices. We introduce \textbf{\emph{ExpoMamba}}, a novel architecture that integrates a frequency-aware state-space model within a modified U-Net. ExpoMamba is designed to address mixed-exposure challenges by \textbf{decoupling the modeling of amplitude (intensity) and phase (structure)} in the frequency domain. This allows for targeted enhancement, making it highly effective for real-time applications, including downstream tasks like object detection and segmentation. Our experiments on six benchmark datasets show that ExpoMamba is up to \textbf{2-3x} faster than competing models and achieves a \textbf{6.8\% PSNR improvement}, establishing a new state-of-the-art in efficient, high-quality low-light enhancement. Source code:~\href{https://github.com/eashanadhikarla/ExpoMamba}{github.com/eashanadhikarla/ExpoMamba}.
\end{abstract}
\vspace{-.1in}
\section{Introduction}
\label{sec:intro}
\vspace{-.05in}

    Enhancing low-light images is crucial for applications ranging from consumer gadgets like phone cameras~\cite{liba2019handheld, liu2024ntire} to sophisticated surveillance systems~\cite{xian2024crose,guo2024hawkdrive,shrivastav2024advancements}. Traditional techniques~\cite{dale1993study,singh2015enhancement,khan2014segment,land1971lightness,ren2020lr3m}
    often struggle to balance processing speed and image quality, particularly with high-resolution, resulting in noise and color distortion in scenarios that demand fast processing such as mobile photography and real-time video streaming.

    \textbf{Limitations of Current Approaches.}~Foundation models have revolutionized computer vision, including low-light image enhancement, by introducing advanced architectures that model complex relationships within image data.
    In particular, transformer-based \cite{wang2023ultra, chen2021pre, zhou2023fourmer,10.1007/978-3-031-78110-0_17} and diffusion-based \cite{wang2023exposurediffusion, wang2023lldiffusion, zhou2023pyramid} low-light techniques have made significant strides. However, the sampling process requires a computationally intensive iterative procedure, and the quadratic runtime of self-attention in transformers makes them unsuitable for real-time use on edge devices where limited processing power and battery constraints pose  challenges. Innovations such as linear attention \cite{katharopoulos-et-al-2020, DBLP:conf/wacv/ShenZZY021, wang2020linformer}, self-attention approximation, windowing, striding \cite{kitaev2020reformer,big-bird}, attention score sparsification \cite{sparse-attention-trans}, hashing \cite{hash-transformer}, and self-attention operation kernelization \cite{katharopoulos-et-al-2020,NEURIPS2021_b1d10e7b,NEURIPS2021_10a7cdd9} 
    have aimed to address these complexities, but often at the cost of increased computation errors compared to simple self-attention \cite{pmlr-v201-duman-keles23a,dosovitskiy2020vit}. (See details in Section~\ref{sec:rw})

    \textbf{Purpose.}~With the rising need for better images, advanced small camera sensors in edge devices have made it more common for customers to capture high-quality images and use them in real-time applications like mobile, laptop, and tablet cameras \cite{morikawa2021image}. However, they all struggle with non-ideal and low-lighting conditions in the real world. Our objective is to achieve a practical balance between perceptual quality (e.g., CIDNet~\cite{feng2024hvi}) and runtime efficiency (e.g., IAT~\cite{Cui_2022_BMVC}, Zero-DCE++~\cite{Zero-DCE++}), while also demonstrating improvements on downstream vision tasks—-such as object detection and segmentation—-that are critical for machine perception in autonomous driving and robotic systems operating under poor lighting.

    \textbf{Contributions.}~Our contributions are summarized as: \textbf{(i)} We introduce the use of Mamba for efficient low-light image enhancement (LLIE), specifically focusing on mixed exposure challenges, where underlit (insufficient brightness) and overlit (excessive brightness) exist in the same image frame.\ \textbf{(ii)} We propose a novel Frequency State Space Block (FSSB) that combines two distinct 2D-Mamba blocks for decoupled modeling of amplitude and phase, allowing the model to enhance brightness and structural consistency separately—-critical for perceptual quality in low-light scenes. \textbf{(iii)} We describe a novel dynamic batch training scheme to improve robustness of multi-resolution inference in our proposed model. \textbf{(iv)} We implement dynamic processing of the amplitude component to highlight distortion (noise, illumination) and the phase component for image smoothing and noise reduction.

\begin{figure*}[!ht]
    \centering
    \includegraphics[width=0.90\textwidth]{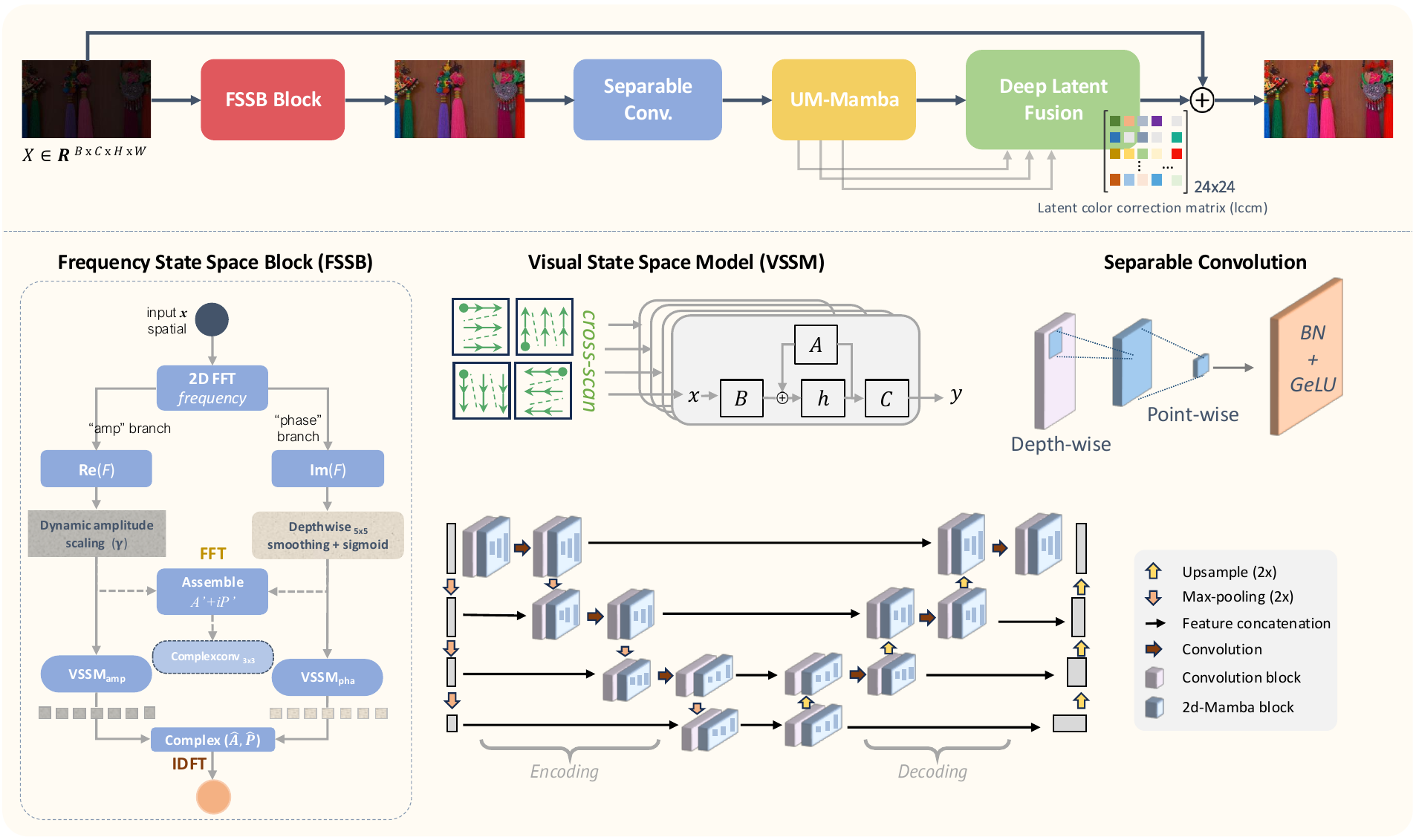}
    \caption{\label{fig:expomamba-arch}\textbf{Overview of the ExpoMamba Architecture.}~ The diagram illustrates the information flow through the \emph{ExpoMamba} model. The architecture efficiently processes sRGB images by integrating convolutional layers, 2D-Mamba blocks, and deep supervision mechanisms to enhance image reconstruction, particularly in low-light conditions.\vspace{-0.10cm}}
\end{figure*}
\section{Related Work}\vspace{-.07in}
\label{sec:rw}

    \subsection{Addressing Performance Bottlenecks}
    The evolution of LLIE has been driven by a cycle of architectural innovation. Although CNNs established the power of data-driven approaches, their inherently local receptive fields struggle to model global illumination inconsistencies. This led to the adoption of Transformers (LLFormer~\cite{wang2023ultra}, IAT~\cite{Cui_2022_BMVC}, IPT~\cite{chen2021pre}, Fourmer~\cite{zhou2023fourmer}, LYT-Net~\cite{brateanu2024lyt}), whose self-attention mechanism excels at capturing long-range dependencies, achieving state-of-the-art results in models like Retinexformer \cite{cai2023retinexformer}. However, this power comes at a steep cost: the ``quadratic bottleneck." The self-attention mechanism scales with a complexity of $O(n^2)$ to image patch count, making Transformers prohibitively slow and memory-intensive for high-resolution imagery and real-time deployment on edge devices.
    
    This fundamental efficiency challenge necessitated a new paradigm. State-Space Models (SSMs), particularly the Mamba architecture \cite{gu2023mamba}, have emerged as the definitive solution. Mamba-based models can capture global context and long-range dependencies with linear-time complexity, $O(n)$, matching the efficiency of RNNs while retaining the parallelizable training of CNNs and the expressive power of Transformers. This breakthrough resolves a long-standing architectural trade-off, paving the way for a new generation of efficient and powerful restoration models.

    \subsection{Mamba-based Low-Light Image Enhancement}
    Mamba's introduction has spurred a new generation of LLIE models. Many are \emph{retinex-inspired frameworks}, including deep unfolding networks like LLEMamba~\cite{zhang2024llemamba}, backbone replacements like RetinexMamba~\cite{bai2025retinexmamba}, artifact-reducing models like MambaLLIE~\cite{weng2024mamballie}, and efficiency-focused variants like EffRetMamba~\cite{EffRetMamba}. Other works explore novel paradigms, from the semantic scanning of BSMamba~\cite{zhang2025bsmamba} to the real-time residual prediction of ResVMUNetX~\cite{wang2024resvmunetx} and the data-efficient learning of Semi-LLIE~\cite{li2024semi}.\ A third group modifies Mamba's interaction space by combining it with frequency or wavelet transforms, as seen in Fourier TMamba \cite{peng2024low}, Wavelet-Mamba~\cite{tan2024wavelet} and WaveMamba~\cite{zou2024wave}.
    
    Despite these innovations, prior works largely treat Mamba as a backbone replacement, a sequence re-ordering pre-processor, or a component within existing spatial frameworks.~\textbf{ExpoMamba}\ innovates at a more fundamental level by operating directly in the frequency domain. Our Frequency State Space Block (FSSB) is the first to employ a dual-branch Mamba architecture to explicitly and independently process amplitude (illumination) and phase (structure). This principled, signal-level decomposition enables fine-grained control, distinguishing our method from those reliant on high-level theories like Retinex or input-reordering schemes. By synergizing the global nature of the Fourier transform with Mamba's linear-time modeling, ExpoMamba establishes a new and effective pathway for LLIE. Additional related work is in \textbf{Appx.\ A}. 

\vspace{-.05in}
\section{Exposure Mamba}
\vspace{-.05in}
    Building upon recent advances in efficient sequence modeling \cite{gu2023mamba,vim,wang2024mambaunet}, we propose {\em ExpoMamba}—a hybrid architecture designed for low-light image enhancement that fuses frequency-aware and spatial learning. The model adopts a modified U-Net structure (combination of $U^2-Net$ \cite{Qin_2020_PR} and M-Net \cite{mehta2017m}), where each encoder-decoder block combines a traditional spatial convolutional unit with a Frequency State Space Block. Each FSSB incorporates two Visual SSM (VSSM) modules, based on the Mamba architecture \cite{gu2023mamba}, which independently process frequency-transformed amplitude and phase features.  More details are deferred to Section~\ref{sec:fssb}; here we emphasize that this hybrid design allows ExpoMamba to efficiently capture multi-scale contextual cues critical for restoring under- and over-exposed regions. The overall pipeline employs a 2D scanning strategy on sRGB images with each block performing operations using a convolutional and encoder-style SSM ($x(t)\in \mathbf{R} \rightarrow y(t) \in \mathbf{R}$)\footnote{A state space model is a type of sequence model that transforms a one-dimensional sequence via an implicit hidden state.}.

    \vspace{-.05in}
    \subsection{Frequency State Space Block (FSSB)}
    \label{sec:fssb}

    The Frequency State Space Block (FSSB) integrates state-space modeling with frequency decomposition to enhance low-light images. Unlike conventional spatial-only models that operate on entangled signal content \cite{zheng2024fdvision,zhou2023fourmer}, FSSB performs a Fourier decomposition of the input into amplitude and phase components, allowing for distinct processing of luminance and structural cues. This separation is critical: amplitude carries information about illumination and contrast, while phase governs edges and object boundaries---features tightly correlated with human visual perception \cite{xiao2004phase,zhou2023fourmer}.
    
    The FSSB (as in Fig.~\ref{fig:expomamba-arch}) initiates its processing by transforming the input image $I$ into the frequency domain. We specifically choose the Fourier transform because its global, harmonically ordered spectral decomposition aligns with Mamba’s architectural strengths in efficiently processing continuous 1D sequences. This allows our model to selectively modulate low-frequency bands, which carry most of the critical illumination and structure information, while handling high-frequency noise—a synergy not as readily achieved with more localized transforms like wavelets.
    %
    %
    \begin{equation}
        \mathbf{F}(u,v) = \iint{\mathbf{I}(x,y) e^{-i2\pi(ux+vy)}dx\cdot dy}
    \end{equation}
    where $\mathbf{F}(u,v)$ denotes the frequency domain representation of the image, and $(u,v)$ are the frequency components corresponding to the spatial coordinates $(x,y)$. This transformation allows for the isolation and manipulation of specific frequency components, which is particularly beneficial for enhancing details and managing noise in low-light images. By decomposing the image into its frequency components, we can selectively enhance high-frequency components to improve edge and detail clarity while suppressing low-frequency components that typically contain noise \cite{Lazzarini2017,zhou2022adaptively}. This selective enhancement and suppression improve the overall image quality.

    The core of the FSSB comprises two 2D-Mamba (Visual-SSM) blocks to process the amplitude and phase components separately. While the VSSM architecture is structurally shared for implementation simplicity, note that the model parameters for the amplitude and phase branches are learned \textbf{independently}. This enables each branch to specialize in its respective task: correcting illumination and noise via the amplitude stream, and preserving fine-grained structural detail via the phase stream. These blocks model the state space transformations as follows:
    {
    \setlength{\jot}{2pt}    
    \begin{align}
        \mathbf{h}[t+1] = \mathbf{A}[t]\cdot\mathbf{h}[t] + \mathbf{B}[t].x[t] \\
        \mathbf{y}[t] = \mathbf{C}[t]\cdot \mathbf{h}[t]
    \end{align}
    }
    Here, $\mathbf{A}[t]$, $\mathbf{B}[t]$, and $\mathbf{C}[t]$ are the state matrices that adapt dynamically based on the input features, and $h[t]$ represents the state vector at time $t$. $\mathbf{y}[t]$ represents processed feature at time $t$, capturing the transformed information from the input features. This dual-pathway setup within the FSSB processes amplitude and phase in parallel.
    
    After processing through each of the VSS blocks, the modified amplitude and phase components are recombined and transformed back to the spatial domain using the inverse Fourier transform:
    \begin{equation}
        \mathbf{\hat{I}}(x,y) = \iint{\mathbf{\hat{F}}(u,v) e^{-i2\pi(ux+vy)}du\cdot dv}
    \end{equation}
    where $\mathbf{\hat{F}}(u,v)$ is the processed frequency domain representation in the latent space of each M-Net block. This method preserves the structural integrity of the image while enhancing textural details that are typically lost in low-light conditions, removing the need of self-attention mechanisms that are widely seen in transformer-based pipelines \cite{10.1145/3530811}. The FSSB also integrates hardware-optimized strategies similar to those employed in the Vision-Mamba architecture \cite{gu2023mamba, vim} such as scan operations and kernel fusion, reducing memory IOs, facilitating efficient data flow between the GPU's memory hierarchies. This optimization significantly reduces computational overhead by a factor of $O(N)$ speeding the operation by $20-40$ times \cite{gu2023mamba}, enhancing processing speed for real-time applications. This can be seen in Fig.~\ref{fig:teaser}, where increasing the resolution size or input length significantly widens the inference time gap, primarily because the multihead attention mechanism in transformer-based models has a computational complexity of $O(N^2)$.

    Within the FSS Block, the amplitude $\mathbf{A}(u, v)$ and phase $\mathbf{P}(u, v)$ components extracted from $\mathbf{F}(u,v)$ are processed through dedicated state-space models. These models, adapted from the Mamba framework, are particularly tailored (dynamic adaptation of state matrices ($\mathbf{A}$, $\mathbf{B}$, $\mathbf{C}$; details in \textbf{Appx. D}) based on spectral properties and the dual processing of amplitude and phase components\footnote{refer to FSSB module in Section~\ref{sec:ablation}.}) to enhance information across frequencies, effectively addressing the typical loss of detail in low-light conditions.

    \paragraph{Amplitude and Phase Component Modeling.} 
    
    Each component $\mathbf{A}(u, v)$ and $\mathbf{P}(u, v)$ undergoes separate but parallel processing paths, modeled by: $s_{t+1} = \mathbf{A}(s_{t}) + \mathbf{BF(x_t)}, y_t = \mathbf{Cs_{t}}$; where, $s_t$ denotes the state at time t, $\mathbf{F(x_t)}$ represents the frequency-domain input at time $t$ (either amplitude or phase), and $\mathbf{A,B,C}$ are the state-space matrices that dynamically adapt during training.

    \subsection{Phase Manipulation}
    \label{sec:phase-proof}
    For an image $I(x, y)$, its Fourier transform $\mathbf{F}(u, v) = \mathbf{A}(u, v) e^{i\phi(u, v)}$, where $\mathbf{A}(u, v)$ is the amplitude and $\phi(u, v)$ is the phase. A uniform shift $\Delta\phi$ to the phase results in a modified image $I'(x, y)$. The derivation is as follows:
    {
    \setlength{\jot}{2pt}    
    \begin{align}
        I'(x, y) &= \iint \mathbf{A}(u, v) e^{i(\phi(u, v) + \Delta\phi)} e^{i2\pi(ux + vy)} \,du\,dv \\
        &= \iint \mathbf{A}(u, v) e^{i\phi(u, v)} e^{i\Delta\phi} e^{i2\pi(ux + vy)} \,du\,dv \\
        &= e^{i\Delta\phi} \iint \mathbf{A}(u, v) e^{i\phi(u, v)} e^{i2\pi(ux + vy)} \,du\,dv \\
        &= (\cos(\Delta\phi) + i\sin(\Delta\phi)) I(x, y)
    \end{align}
    }
    This formulation reveals that a uniform phase shift transforms the spatial structure of the image. As prior studies show \cite{xiao2004phase}, human vision is highly sensitive to phase variations. By adaptively modifying the phase, our model preserves and refines scene structure (e.g., edges, contours), while the amplitude branch handles illumination and contrast. We use phase offset as a single scalar per FSSB stage, denoted $\Delta\phi_{s}$, learned during training and applied uniformly to all $(u,v)$. For stability it is bounded to $[-\Delta\phi_{\max},\,\Delta\phi_{\max}]$ with a smooth clipping (extended details in \textbf{Appx. B}). 

    \subsection{Algorithm}
    
    \noindent The pseudocode found in Algorithm~\ref{sec:algorithm} presents the details of ExpoMamba training with FSSB blocks.

    \begin{algorithm}[!ht]
    \small
    \caption{Concise ExpoMamba Training w/ FSSB}\label{sec:algorithm}
    \begin{threeparttable}
    \begin{algorithmic}[1]
    \REQUIRE dataset $\mathcal{D}$,\; epochs $E$,\; model params $\theta$ (FSSB $\rightarrow$ HDR $\rightarrow$ ComplexConv)
    \FOR{$I\!\in\!\mathcal{D}$}                                      
        \STATE $(A,P)\leftarrow\mathrm{FFT}(I)$
        \STATE $(A'',P'')\leftarrow\mathrm{VSSM}(A,P)$
        \STATE $\hat I\leftarrow\mathrm{IFFT}(A''+iP'')$
    \ENDFOR
    \FOR{$e=1$ \textbf{to} $E$}                                         
        \FOR{mini-batch $\mathcal{B}$}
            \STATE $\hat{\mathcal{B}}\leftarrow\text{ComplexConv}(\text{HDR}(\text{FSSB}(\mathcal{B})))$
            \STATE $\theta\leftarrow\theta-\eta\nabla_{\theta}\mathcal{L}(\hat{\mathcal{B}},\mathcal{B})$
        \ENDFOR
    \ENDFOR
    \RETURN $\theta$
    \end{algorithmic}
        \begin{tablenotes} 
            \footnotesize \item A detailed version can be found in Supp.~\ref{alg:full}.
        \end{tablenotes}
    \end{threeparttable}
    \end{algorithm}

    \paragraph{Latent Space Color Correction Matrix (LCCM)}
    \textbf{Why latent instead of image space.}
    \emph{\textbf{(1)} Precision.} Latent tensors are floating point, while image space is typically 8 to 10 bit per channel. Applying color mixing in latent space avoids rounding and clipping, so small corrections can be learned and accumulated without staircasing. \emph{\textbf{(2)} Expressivity.} We act on a $C$\,channel feature basis rather than only RGB. This enables a $C$$\times$$C$ CCM (we use $C{=}24$) that generalizes the standard $3\times 3$ matrix and can model sensor and ISP shifts as well as scene dependent tints and cross-channel couplings. \emph{\textbf{(3)} Semantics and stability.} Latent features carry edges, textures, and mixed exposure cues that are attenuated in pixels.
    Operating in this space makes the correction more stable under low light noise and uneven illumination.
    
    \textbf{Formulation and placement.}
    Given $X\!\in\!\mathbb{R}^{C\times H\times W}$, the LCCM is a per-pixel linear map $W\!\in\!\mathbb{R}^{C\times C}$ with bias $b$,\vspace{-.1in}%
    \begin{align}
        Y(:,h,w)=W\,X(:,h,w)+b,
    \end{align}
    implemented as a $1{\times}1$ convolution.
    We initialize $W{=}I_C$, $b{=}0$, and add a small penalty $\lambda\lVert W{-}I_C\rVert_F^2$ to discourage unnecessary drift. We place one LCCM before HDR gating in each decoder FSSB and a final LCCM before the output head. The cost is $O(C^2HW)$; with $C{=}24$ the per-pixel multiplies are $576$, which is negligible relative to surrounding blocks.
    
    As noted in ablation study~\ref{tab:ablation}, the LCCM corrects sensor and ISP color shifts while reducing banding and quantization artifacts, aligning with ExpoMamba’s goal of high fidelity without relying on post-hoc RGB fixes.

    \vspace{-.1in}
    \paragraph{Feature Recovery in FSSB.}

\begin{figure}[tb]
        \centering
        \includegraphics[width=0.49\textwidth]{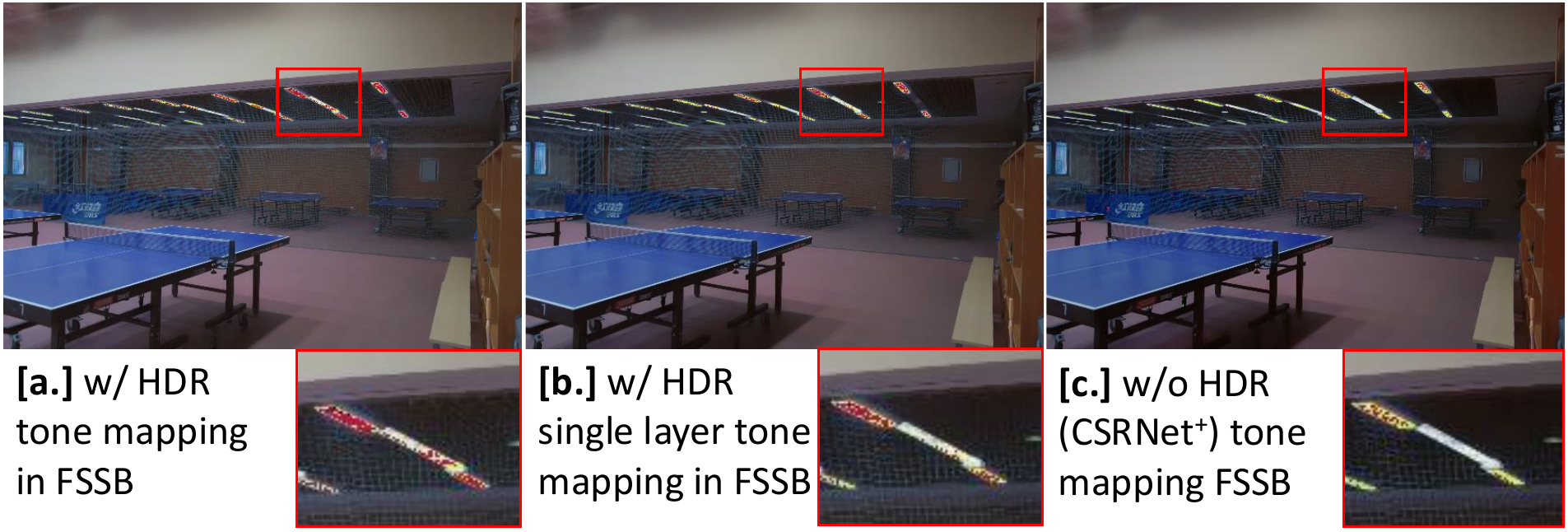}
        \caption{\label{fig:tone_mapping}The effectiveness of various HDR tone mapping layers inside the FSS block. $\text{CSRNet}^{+}$ with shrunken conditional blocks and dilated convolutions lessens overexposed artifacts.\vspace{-0.05in}}
    \end{figure}

    We implement an HDR gate in feature space. Overexposed activations are detected with a soft mask $M=\sigma\!\big(k\,(Y-\tau)\big)$ and corrected using a gentle tone curve
    $T(x)=\log(1+s\,x)$ or a learned CSRNet variant. We blend only where $M$ is high:
    $\tilde{x}=(1-M)\circ x + M\circ T(x)$. This layer sits at the end of each FSSB and before the final output head, which attenuates halos and blooming while preserving detail. Unless stated otherwise we use $\tau=0.90$. Figure~\ref{fig:tone_mapping} compares the parametric and CSRNet$^{+}$ options.
    
    We leverage the ComplexConv function from complex networks as introduced by Trabelsi et al.~\cite{TrabelsiBZSSSMR18}. This function is incorporated into our model to capture and process additional information beyond traditional real-valued convolutions. Specifically, the ComplexConv function allows the simultaneous manipulation of the amplitude and phase information in the frequency domain, which is essential to preserve the integrity of textural details in low-light images. The dual processing of amplitude and phase ensures that each component is optimized separately. Tone mapping and ComplexConv have proven effective in overcoming limitations of traditional image processing techniques \cite{9707563, liu2024design}. We integrate these methods into our FSS design to address adverse lighting conditions in low-light environments.
    The input components in the frequency representation are processed through dynamic amplitude scaling and phase continuity layer as shown in  Fig.~\ref{fig:expomamba-arch}. As claimed by Fourmer~\cite{zhou2023fourmer}, we have determined that the primary source of image degradation is indeed amplitude, specifically in the area between the amplitude and phase division within the image. Moreover, we found the amplitude component primarily affects brightness of the image, directly impacting the visibility and the sharpness of the features within the image. However, the phase component encodes the positional information of these features, defining the structure and the layout of the image. The phase component of the image has previously been found to have a close relation with perceptual analysis \cite{xiao2004phase}. Along those lines, we show that the human visual system is more sensitive to changes in phase rather than amplitude (see Section~\ref{sec:phase-proof}).

\begin{table}[t]
    \centering
    \caption{\label{tab:lolv1-lolv2}Comparing four popular metrics such that every column showcases the top three methods; $\firsttone{dark}$, $\secondtone{medium}$, and $\thirdtone{light}$ shades of blue representing the best, second best, and third best models among the proposed and all popular SOTA models.}\vspace{-0.2cm}
    \begin{threeparttable}
    \resizebox{\columnwidth}{!}{ 
    \begin{tabular}{ccccccccccc}
    \toprule
        \multirow{2}{*}{\textbf{Methods}} &
        \multicolumn{3}{c}{\textbf{LOLv1}} &
        &
        \multicolumn{3}{c}{\textbf{LOLv2 (Real Captured)}} &
        \multirow{2}{*}{\begin{tabular}[c]{@{}c@{}}\textbf{Inference}\\\textbf{time (ms)}\end{tabular}} \\ \cmidrule{2-8}
        & \textbf{PSNR} $\uparrow$ & \textbf{SSIM} $\uparrow$ & \textbf{LPIPS} $\downarrow$ & 
        & 
        \textbf{PSNR} $\uparrow$ & \textbf{SSIM} $\uparrow$ & \textbf{LPIPS} $\downarrow$ & 
        & \\ \midrule
        \multicolumn{1}{c}{Retinex$^{\ddag}$} &
        16.774 & 0.462 & 0.417 & 
        &
        17.715 & 0.652  & 0.436  & 
        \multicolumn{1}{c}{4493} \\
        \multicolumn{1}{c}{MIRNet} &
        \thirdtone{24.138} & 0.830 & 0.250 &
        &
        20.020 & 0.82 & 0.233 &
        \multicolumn{1}{c}{1795} \\
        \multicolumn{1}{c}{EnlightenGAN$^{\ddag}$} &
        17.606 & 0.653 & 0.372 & 
        &
        18.676 & 0.678 & 0.364 & 
        \multicolumn{1}{c}{-} \\
        \multicolumn{1}{c}{ReLLIE$^{\ddag}$} &
        11.437 & 0.482 & 0.375 & 
        &
        14.400 & 0.536 & 0.334 & 
        \multicolumn{1}{c}{\firsttone{3.500}} \\
        \multicolumn{1}{c}{RUAS$^{\ddag}$} &
        16.405 & 0.503 & 0.364 & 
        &
        15.351 & 0.495 & 0.395 & 
        \multicolumn{1}{c}{\secondtone{15.51}} \\
        \multicolumn{1}{c}{DDIM} &
        16.521 & 0.776 & 0.376 & 
        &
        15.280 & 0.788  & 0.387  & 
        \multicolumn{1}{c}{1213} \\
        \multicolumn{1}{c}{CDEF} &
        16.335 & 0.585 & 0.407 & 
        &
        19.757 & 0.63   & 0.349  & 
        \multicolumn{1}{c}{-} \\
        \multicolumn{1}{c}{SCI} &
        14.784 & 0.525 & 0.366 & 
        &
        17.304 & 0.54 & 0.345 & 
        \multicolumn{1}{c}{1755} \\
        \multicolumn{1}{c}{URetinex-Net} &
        19.842 & 0.824 & 0.237 & 
        &
        21.093 & 0.858  & 0.208  
        & \multicolumn{1}{c}{1804} \\
        \multicolumn{1}{c}{SNRNet$^{\ddag}$} &
        23.432 & 0.843 & 0.234 & 
        &
        21.480 & 0.849 & 0.237 & 
        \multicolumn{1}{c}{72.16} \\
        \multicolumn{1}{c}{Uformer$^\star$} &
        19.001 & 0.741 & 0.354 & 
        &
        18.442 & 0.759 & 0.347 & 
        \multicolumn{1}{c}{901.2} \\
        \multicolumn{1}{c}{Restormer$^\star$} &
        20.614 & 0.797 & 0.288 & 
        &
        24.910 & 0.851 & 0.264 & 
        \multicolumn{1}{c}{513.1} \\
        \multicolumn{1}{c}{Palette$^\clubsuit$} &
        11.771 & 0.561 & 0.498 & 
        &
        14.703 & 0.692 & 0.333 & 
        \multicolumn{1}{c}{168.5} \\
        \multicolumn{1}{c}{UHDFour$^\ddag$} &
        23.093 & 0.821 & 0.259 & 
        &
        21.785 & 0.854 & 0.292 & 
        \multicolumn{1}{c}{64.92} \\
        \multicolumn{1}{c}{WeatherDiff$^\clubsuit$} &
        17.913 & 0.811 & 0.272 & 
        &
        20.009 & 0.829 & 0.253 & 
        \multicolumn{1}{c}{5271} \\
        \multicolumn{1}{c}{GDP$^\clubsuit$} &
        15.896 & 0.542 & 0.421 & 
        &
        14.290 & 0.493 & 0.435 & 
        \multicolumn{1}{c}{-} \\
        \multicolumn{1}{c}{DiffLL$^\clubsuit$} &
        \firsttone{26.336} & 0.845 & 0.217 & 
        &
        \firsttone{28.857} & \secondtone{0.876} & 0.207 & 
        \multicolumn{1}{c}{157.9} \\
        \multicolumn{1}{c}{CIDNet$^\ddag$} &
        23.090 & \secondtone{0.851} & 0.085 & 
        &
        \thirdtone{23.220} & 0.863 & \firsttone{0.103} & 
        \multicolumn{1}{c}{-} \\
        \multicolumn{1}{c}{LLformer$^\star$} &
        22.890 & 0.816 & \secondtone{0.202} & 
        &
        23.128 & 0.855 & 0.153 & 
        \multicolumn{1}{c}{1956} \\ \midrule
        \multicolumn{1}{c}{\textbf{ExpoMamba}} 
        & 
        22.870 & 0.845 & 0.215 & 
        &
        23.000 & 0.860 & 0.203 & 
        \multicolumn{1}{c}{\thirdtone{36.00}} \\
        \multicolumn{1}{c}{$\textbf{ExpoMamba}_{da}$} &
        23.092 & \thirdtone{0.847} & 0.214 & 
        &
        23.131 & \thirdtone{0.868} & 0.224 & 
        \multicolumn{1}{c}{38.00} \\
        \multicolumn{1}{c}{$\textbf{ExpoMamba}_{gt}$} 
        & 
        \secondtone{25.770} & \firsttone{0.860} & \thirdtone{0.212} & 
        &
        \secondtone{28.040} & \firsttone{0.885} & 0.232 & 
        \multicolumn{1}{c}{\thirdtone{36.00}} \\ \bottomrule
    \end{tabular}
    }
    
    \begin{tablenotes} 
        \tiny
        \item ``\textbf{da}'' - Dynamic adjustment. (refer Section-\ref{sec:ablation}) ~/~ ``\textbf{gt}'' - With ground-truth mean.
    \end{tablenotes}
    \end{threeparttable}
    \vspace{-.05in}
\end{table}

    \vspace{-.03in}
    \subsection{Multi-modal Feature Learning} \vspace{-.04in}

    The inherent complexity of low-light images, where both underexposed and overexposed elements coexist, necessitates a versatile approach to image processing. Traditional methods, which typically focus either on spatial details or frequency-based features, fail to adequately address the full spectrum of distortions encountered in such environments. By contrast, the hybrid modeling approach of ExpoMamba leverages the strengths of both the spatial and frequency domains, facilitating a more comprehensive and nuanced enhancement of image quality.

    Operations like the Fourier transform offer the ability to isolate distortion modes by frequency, allowing ExpoMamba to denoise brightness artifacts via amplitude manipulation and preserve fine-grained structures via phase-aware reconstruction. This domain provides a global view of the image data, allowing for the manipulation of features that are not easily discernible in the spatial layout. Simultaneously, the spatial domain is critical for maintaining the local coherence of image features, ensuring that enhancements do not introduce unnatural artifacts. Finally, the hybrid-modeled features pass through deep supervision, where we combine ExpoMamba’s intermediate layer outputs, apply a color correction matrix (CCM) in the latent dimensions during deep supervision, and pass through the final layer.

    While most LLIE approaches focus solely on perceptual quality, ExpoMamba is explicitly evaluated on downstream tasks relevant to autonomous systems, such as object detection and scene parsing. Our frequency-aware hybrid design preserves structural integrity and suppresses noise, which are essential to improve downstream model performance (e.g., YOLOv3 and DeepLab-V3+) on low-light images. This bridges the gap between low-level enhancement and high-level vision reliability in real-world applications.
\vspace{-.03in}
\section{Experiments and Implementation Details}
\vspace{-.04in}

    We evaluate ExpoMamba through a series of experiments. We begin by outlining the datasets used, experimental setup, followed by a comparison of our method against state-of-the-art techniques using four quantitative metrics. We also perform a detailed ablation study in Tab.~\ref{tab:ablation}/Sec.~\ref{sec:ablation}.

    \vspace{-.025in}
    \subsection{Datasets}
    \vspace{-.025in}

    We tested the efficacy of our model on four datasets: \textbf{(1) LOL}~\cite{WeiWY018}. LOLv2 \cite{Yang_2020_CVPR} is divided into real and synthetic subsets. The training and testing sets are split into 485/15, 689/100, and 900/100 on LOLv1, LOLv2-real, and LOLv2-synthetic with $3\times 400\times 600$ resolution images. \textbf{(2) LOL4K} is an ultra-high definition dataset with $3\times 3840\times 2160$ resolution images, containing 8,099 pairs of low-light/normal-light images, split into 5,999 pairs for training and 2,100 pairs for testing. \textbf{(3) SICE} \cite{Cai2018deep} includes 4,800 images,  real and synthetic, at various exposure levels and resolutions, divided into training, validation, and testing sets in a 7:1:2 ratio. \textbf{(4) ExDark}~\cite{Exdark} is a real-world object detection dataset comprising 7,363 low-light images from 12 object categories, with annotated bounding boxes. We follow an 80/20 training/test split and resize images to $608 \times 608$ resolution for YOLOv3-based detection. \textbf{(5) ACDC}~\cite{SDV21} is a nighttime urban scene segmentation dataset containing 1,006 images. We fine-tune DeepLab-V3+ on its night-time training subset and evaluate on the validation set using mIoU (mean Intersection over Union).

\begin{table}[!t]
    \centering
    \begin{center}
    \caption{\label{tab:4k}Evaluation on the UHD-LOL4K dataset. $\dagger, \ddag, \S, \triangle$, and ${\star}$ denote supervised CNN, unsupervised CNN, zero-shot, and transformer-based models, respectively. We use dynamic adjustment for both `s' and `l' ExpoMamba models during inference.\vspace{-0.1cm}}
    \scalebox{0.68}{
        \begin{tabular}{lcccc} \toprule
        \multirow{2}{*}{Methods} & \multicolumn{4}{c}{ \textbf{UHD-LOL4K}} \\ \cmidrule(r){2-5}
            & PSNR $\uparrow$ & SSIM $\uparrow$ & LPIPS $\downarrow$ & MAE $\downarrow$ \\ \midrule

            SRIE$^{\dagger}$~\cite{fu2016weighted} & 16.7730 & 0.8365 & 0.1495 & 0.1416 \\
            MSRCR$^{\dagger}$~\cite{jobson1997multiscale} & 12.5238 & 0.8106 & 0.2136 & 0.2039 \\
            RetinexNet$^{\ddag}$~\cite{WeiWY018} & 21.6702 & 0.9086 & 0.1478& 0.0690 \\
            DSLR$^{\ddag}$~\cite{lim2020dslr} & 27.3361 & 0.9231 & 0.1217 & 0.0341 \\
            KinD$^{\ddag}$~\cite{zhang2019kindling} & 18.4638 & 0.8863 & 0.1297 & 0.1060 \\
            Z\_DCE$^{\S}$~\cite{guo2020zero} & 17.1873 & 0.8498 & 0.1925 & 0.1465 \\
            Z\_DCE++$^{\S}$~\cite{Zero-DCE++} & 15.5793 & 0.8346 & 0.2223 & 0.1701 \\
            RUAS$^{\triangle}$~\cite{liu2021retinex} & 14.6806 & 0.7575 & 0.2736 & 0.1690 \\
            ELGAN$^{\triangle}$~\cite{jiang2021enlightengan} & 18.3693 & 0.8642 & 0.1967 & 0.1011 \\
            Uformer${\star}$~\cite{wang2021uformer} & 29.9870 & 0.9804 & 0.0342 & 0.0376 \\
            Restormer${\star}$~\cite{zamir2021restormer} & \secondtone{36.9094} & 0.9881 & \thirdtone{0.0226} & \secondtone{0.0117} \\
            LLFormer${\star}$~\cite{wang2023ultra} & \firsttone{37.3340} & \thirdtone{0.9862} & \firsttone{0.0200} & \firsttone{0.0116} \\
            UHD-Four~\cite{li2023embedding} & 35.1010 & \firsttone{0.9901} & \secondtone{0.0210} & - \\ \bottomrule
            \textbf{$\text{ExpoMamba}_{s}$} & 28.3300 & 0.9730 & 0.0820 & \thirdtone{0.0315} \\
            \textbf{$\text{ExpoMamba}_{l}$} & \thirdtone{35.2300} & \secondtone{0.9890} & 0.0630 & 0.0451 \\ \bottomrule 
        \end{tabular}
        }
    \end{center}
    \vspace{-0.35cm}
\end{table}

\begin{table}[t]
\centering
\caption{\label{tab:mamba_comparison}Performance comparison with recent Mamba-based LLIE models on LOLv1 and LOLv2 datasets.\vspace{-0.25cm}}
\scalebox{0.65}{
        \begin{tabular}{lcccc}
        \toprule
        \multirow{2}{*}{\textbf{Model}} & \multicolumn{2}{c}{\textbf{LOLv1}} & \multicolumn{2}{c}{\textbf{LOLv2}} \\
        \cmidrule{2-5} & 
        \textbf{PSNR} $\uparrow$ & \textbf{SSIM} $\uparrow$ & \textbf{PSNR} $\uparrow$ & \textbf{SSIM} $\uparrow$ \\ \midrule
        
        RetinexMamba~\cite{bai2025retinexmamba} & \thirdtone{24.02} & 0.827 & 22.45 & 0.844 \\
        Fourier TMamba~\cite{peng2024low}       & \secondtone{24.33} & 0.845 & 22.41 & 0.860 \\
        Wavelet-Mamba~\cite{tan2024wavelet}     & 23.27 & 0.851 & 22.49 & 0.869 \\
        MambaLLIE~\cite{weng2024mamballie}      & -     & -     & 22.95 & 0.847 \\
        EffRetMamba~\cite{EffRetMamba}          & 23.65 & 0.837 & 22.32 & 0.841 \\ \midrule
        \textbf{ExpoMamba\textsubscript{gt}} & \firsttone{\textbf{25.77}} & \firsttone{\textbf{0.860}} & \textbf{28.04} & \textbf{0.885} \\ \bottomrule
    \end{tabular}
}
\end{table}

    \vspace{-.025in}
    \subsection{Experimental setting}
    \vspace{-.02in}
    
    \begin{figure*}[!ht]
    \centering
    \begin{minipage}{0.52\textwidth}
        \centering
        \includegraphics[width=\linewidth]{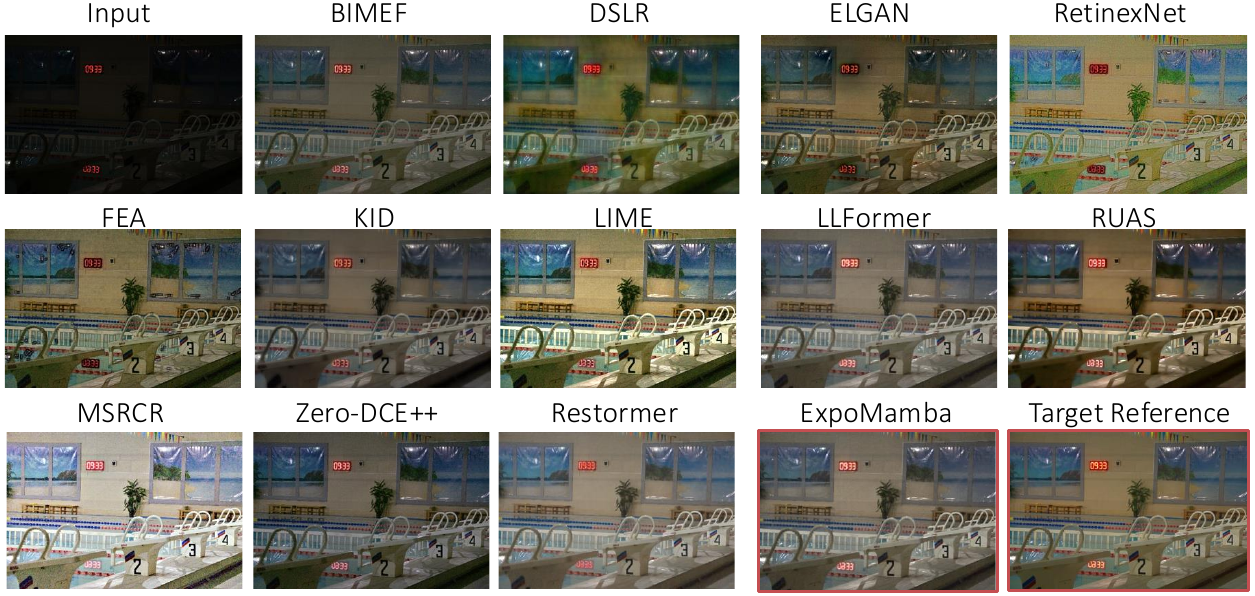}\vspace{-.05in}
        \caption{\label{fig:lolv1b}Qualitative comparison of ExpoMamba and baselines on the LOLv1 dataset. Results demonstrate structural fidelity and color balance under mixed lighting conditions.}
    \end{minipage}%
    \hfill
    \begin{minipage}{0.45\textwidth}
        \centering
        \captionof{table}{\label{tab:downstream-tasks}Experimental results on low-light object detection (ExDark) and segmentation (ACDC) tasks.}\vspace{-0.25cm}
        \scalebox{0.75}{
        \begin{tabular}{ccccc}
            \toprule
            \multirow{2}{*}{\textbf{Methods}} & 
            \multicolumn{2}{c}{\textbf{ExDark Detection}} & 
            \multicolumn{2}{c}{\textbf{ACDC Segmentation}} \\ 
            \cmidrule{2-5} 
            & mAP~$\uparrow$ & time (s)~$\downarrow$ & mIOU~$\uparrow$ & time (s)~$\downarrow$ \\ \midrule
            
            YOLOv3 \cite{yolov3}           & 76.4 & \firsttone{0.033} & 63.3 & \firsttone{0.249} \\
            MBLLEN \cite{Lv2018MBLLEN}     & 76.3 & 0.086 & \thirdtone{63.0} & 0.332 \\
            DeepLPF \cite{Moran2020}       & 76.3 & 0.138 & 61.9 & 0.807 \\
            Zero-DCE \cite{Zero-DCE}       & 76.9 & \thirdtone{0.042} & 61.9 & 0.300 \\
            MAET (w ort) \cite{MAET}       & 74.0 & 0.123 & - & - \\
            IAT \cite{Cui_2022_BMVC}       & \thirdtone{77.2} & \secondtone{0.040} & 62.1 & \secondtone{0.280} \\
            ILE-YOLO \cite{WANG2025126504} & \secondtone{78.5} & 0.047 & \secondtone{64.3} & 0.306 \\ \midrule
            \textbf{ExpoMamba}                      & \firsttone{79.8} & 0.076 & \firsttone{64.3} & \thirdtone{0.297} \\
            \bottomrule
        \end{tabular}}
    \end{minipage}
    \vspace{-.1in}
\end{figure*}

    The proposed network is a single-stage end-to-end training model. The patch sizes are set to $128 \times 128$, $256 \times 256$, and $324 \times 324$ with checkpoint restarts and batch sizes of $8$, $6$, and $4$, respectively, in consecutive runs. For dynamic patch training, we use different patch sizes simultaneously. To improve robustness across input resolutions, we adopt a dynamic patch training strategy randomizing patch sizes across batches. This enhances scale-invariance and supports efficient multi-resolution inference (more details in \textbf{Appx.\ E}). The optimizer is \texttt{RMSProp} with learning rate of $1$$\times$$10^{-4}$, weight decay of $1$$\times$$10^{-7}$, and momentum of $0.9$. A linear warm-up cosine annealing 
    scheduler with $15$ warm-up epochs is used, starting with a learning rate of $1$$\times$$10^{-4}$. All experiments are conducted on NVIDIA A10 GPU. For model configuration details see Tab.~\ref{tab:model-config}.

    For downstream tasks, we follow the configurations used in IAT~\cite{Cui_2022_BMVC} and ILE-YOLO~\cite{WANG2025126504}. For object detection, we train $608\times608$ with an 80/20 split using SGD (momentum=0.9, weight decay=$1$$\times$$10^{-4}$, learning rate=$1$$\times$$10^{-3}$, batch size=8). For segmentation, DeepLab-V3+ (ResNet-101 backbone, pre-trained on Cityscapes) is fine-tuned on ACDC-night training images using SGD (initial learning rate=0.05, weight decay=$5$$\times$$10^{-4}$). Performance was collected on raw and enhanced inputs using mean Average Precision (mAP) and mIoU.

    \vspace{-.15in}
    \paragraph{Baselines.} We use official code for MIRNet, Restormer, LLFormer, URetinex-Net, and DiffLL. MIRNet, Restormer, and LLFormer are retrained on LOL-v1/v2 and SICE with their recommended losses and schedulers. URetinex-Net and DiffLL use released weights due to training cost. When retraining, we adopt each paper’s epochs, augmentations, and loss composition; if absent, we train for 300 epochs with cosine decay and early stop on PSNR.
    

\begin{table}[t]
\centering
\caption{\label{tab:ablation}Ablation of ExpoMamba FSSB components.\vspace{-.1in}}
\resizebox{\columnwidth}{!}{
\begin{tabular}{lcccccc}
    \toprule
    Config & FSSB & HDR & CSRNet$^{+}$ & HDR$_\text{out}$ & DA & PSNR / SSIM \\
    \midrule
    Conv baseline & \xmark & \xmark & \xmark & \xmark & \xmark & 18.978 / 0.815 \\
    + FSSB only & \cmark & \xmark & \xmark & \xmark & \xmark & 19.787 / 0.828 \hspace{0.4em}{\small(+0.809 / +0.013)} \\
    + HDR only & \xmark & \cmark & \xmark & \xmark & \xmark & 22.459 / 0.836 \hspace{0.4em}{\small(+3.481 / +0.021)} \\
    + CSRNet$^{+}$ only & \xmark & \xmark & \cmark & \xmark & \xmark & 20.576 / 0.823 \hspace{0.4em}{\small(+1.598 / +0.008)} \\
    FSSB + HDR & \cmark & \cmark & \xmark & \xmark & \xmark & 24.878 / 0.841 \hspace{0.4em}{\small(+5.900 / +0.026)} \\
    FSSB + HDR + DA + CSRNet$^{+}$ & \cmark & \cmark & \cmark & \xmark & \cmark & 25.110 / 0.845 \hspace{0.4em}{\small(+6.132 / +0.030)} \\
    FSSB + HDR + HDR$_\text{out}$ + DA & \cmark & \cmark & \xmark & \cmark & \cmark & \textbf{25.640} / \textbf{0.860} \hspace{0.4em}{\small(+6.662 / +0.045)} \\
\bottomrule
\end{tabular}}
\vspace{-0.3em}
\end{table}

\begin{table}[!tb]
\begin{center}
    \caption{\label{tab:sicev2} Comparative performance over SICE-v2 \cite{Cai2018deep} datasets. Our `s': smallest model outperforms all the baselines.}\vspace{-0.25cm}
    \scalebox{0.5}{ 
    \begin{tabular}{llllllll}
    \toprule
    \multicolumn{1}{c}{\multirow{3}{*}{\textbf{Method}}} &
    \multicolumn{6}{c}{\textbf{SICE-v2}} &
    \multicolumn{1}{c}{\multirow{3}{*}{\textbf{\#params}}} \\ \cmidrule{2-7}
    \multicolumn{1}{c}{} &
    \multicolumn{2}{c}{\textbf{Underexposure}} &
    \multicolumn{2}{c}{\textbf{Overexposure}} &
    \multicolumn{2}{c}{\textbf{Average}} &
    \multicolumn{1}{c}{} \\ \cmidrule{2-7}
    \multicolumn{1}{c}{} &
    \multicolumn{1}{c}{\textbf{PSNR} $\uparrow$} &
    \multicolumn{1}{c}{\textbf{SSIM} $\uparrow$} &
    \multicolumn{1}{c}{\textbf{PSNR} $\uparrow$} &
    \multicolumn{1}{c}{\textbf{SSIM} $\uparrow$} &
    \multicolumn{1}{c}{\textbf{PSNR} $\uparrow$} &
    \multicolumn{1}{c}{\textbf{SSIM} $\uparrow$} &
    \multicolumn{1}{c}{} \\ \midrule



    RetinexNet \cite{WeiWY018} &
        \multicolumn{1}{c}{12.94} &
        \multicolumn{1}{c}{0.5171} &
        \multicolumn{1}{c}{12.87} &
        \multicolumn{1}{c}{0.5252} &
        \multicolumn{1}{c}{12.90} &
        \multicolumn{1}{c}{0.5212} &
        \multicolumn{1}{c}{0.84M} \\

    URetinexNet \cite{9879970} &
        \multicolumn{1}{c}{12.39} &
        \multicolumn{1}{c}{0.5444} &
        \multicolumn{1}{c}{7.40} &
        \multicolumn{1}{c}{0.4543} &
        \multicolumn{1}{c}{12.40} &
        \multicolumn{1}{c}{0.5496} &
        \multicolumn{1}{c}{1.32M} \\





    DeepUPE \cite{Wang_2019_CVPR} &
        \multicolumn{1}{c}{16.21} &
        \multicolumn{1}{c}{0.6807} &
        \multicolumn{1}{c}{11.98} &
        \multicolumn{1}{c}{0.5967} &
        \multicolumn{1}{c}{14.10} &
        \multicolumn{1}{c}{0.6387} &
        \multicolumn{1}{c}{7.79M} \\



    SID-L \cite{huang2022exposure} &
        \multicolumn{1}{c}{19.43} &
        \multicolumn{1}{c}{0.6644} &
        \multicolumn{1}{c}{17.00} &
        \multicolumn{1}{c}{0.6495} &
        \multicolumn{1}{c}{18.22} &
        \multicolumn{1}{c}{0.6570} &
        \multicolumn{1}{c}{11.56M} \\

    RUAS \cite{ll_benchmark} &
        \multicolumn{1}{c}{16.63} &
        \multicolumn{1}{c}{0.5589} &
        \multicolumn{1}{c}{4.54} &
        \multicolumn{1}{c}{0.3196} &
        \multicolumn{1}{c}{10.59} &
        \multicolumn{1}{c}{0.4394} &
        \multicolumn{1}{c}{0.0014M} \\

    SCI \cite{Ma_2022_CVPR} &
        \multicolumn{1}{c}{17.86} &
        \multicolumn{1}{c}{0.6401} &
        \multicolumn{1}{c}{4.45} &
        \multicolumn{1}{c}{0.3629} &
        \multicolumn{1}{c}{12.49} &
        \multicolumn{1}{c}{0.5051} &
        \multicolumn{1}{c}{0.0003M} \\

    MSEC \cite{afifi2021learning} &
        \multicolumn{1}{c}{19.62} &
        \multicolumn{1}{c}{0.6512} &
        \multicolumn{1}{c}{17.59} &
        \multicolumn{1}{c}{0.6560} &
        \multicolumn{1}{c}{18.58} &
        \multicolumn{1}{c}{0.6536} &
        \multicolumn{1}{c}{7.04M} \\


    LCDPNet \cite{wang2022local} &
        \multicolumn{1}{c}{17.45} &
        \multicolumn{1}{c}{0.5622} &
        \multicolumn{1}{c}{17.04} &
        \multicolumn{1}{c}{0.6463} &
        \multicolumn{1}{c}{17.25} &
        \multicolumn{1}{c}{0.6043} &
        \multicolumn{1}{c}{0.96M} \\

    DRBN \cite{drbn} &
        \multicolumn{1}{c}{17.96} &
        \multicolumn{1}{c}{0.6767} &
        \multicolumn{1}{c}{17.33} &
        \multicolumn{1}{c}{0.6828} &
        \multicolumn{1}{c}{17.65} &
        \multicolumn{1}{c}{0.6798} &
        \multicolumn{1}{c}{0.53M} \\


    DRBN-ERL+ENC \cite{huang2023learning} &
        \multicolumn{1}{c}{\thirdtone{22.06}} &
        \multicolumn{1}{c}{\secondtone{0.7053}} &
        \multicolumn{1}{c}{19.50} &
        \multicolumn{1}{c}{\secondtone{0.7205}} &
        \multicolumn{1}{c}{20.78} &
        \multicolumn{1}{c}{\secondtone{0.7129}} &
        \multicolumn{1}{c}{0.58M} \\

    ELCNet \cite{huang2017arbitrary} &
        \multicolumn{1}{c}{22.05} &
        \multicolumn{1}{c}{0.6893} &
        \multicolumn{1}{c}{19.25} &
        \multicolumn{1}{c}{0.6872} &
        \multicolumn{1}{c}{20.65} &
        \multicolumn{1}{c}{0.6861} &
        \multicolumn{1}{c}{0.018M} \\

    ELCNet+ERL \cite{huang2023learning} &
        \multicolumn{1}{c}{22.14} &
        \multicolumn{1}{c}{\thirdtone{0.6908}} &
        \multicolumn{1}{c}{19.47} &
        \multicolumn{1}{c}{\thirdtone{0.6982}} &
        \multicolumn{1}{c}{20.81} &
        \multicolumn{1}{c}{\thirdtone{0.6945}} &
        \multicolumn{1}{c}{0.018M} \\

    FECNet \cite{10.1145/3343031.3350855} &
        \multicolumn{1}{c}{22.01} &
        \multicolumn{1}{c}{0.6737} &
        \multicolumn{1}{c}{19.91} &
        \multicolumn{1}{c}{0.6961} &
        \multicolumn{1}{c}{20.96} &
        \multicolumn{1}{c}{0.6849} &
        \multicolumn{1}{c}{0.15M} \\

    FECNet+ERL \cite{huang2023learning} &
        \multicolumn{1}{c}{\secondtone{22.35}} &
        \multicolumn{1}{c}{0.6671} &
        \multicolumn{1}{c}{\thirdtone{20.10}} &
        \multicolumn{1}{c}{0.6891} &
        \multicolumn{1}{c}{\thirdtone{21.22}} &
        \multicolumn{1}{c}{0.6781} &
        \multicolumn{1}{c}{0.15M} \\

    IAT \cite{Cui_2022_BMVC} &
        \multicolumn{1}{c}{21.41} &
        \multicolumn{1}{c}{0.6601} &
        \multicolumn{1}{c}{\firsttone{22.29}} &
        \multicolumn{1}{c}{0.6813} &
        \multicolumn{1}{c}{\firsttone{21.85}} &
        \multicolumn{1}{c}{0.6707} &
        \multicolumn{1}{c}{0.090M} \\ \midrule

    $\textbf{ExpoMamba}_{s}$ &
        \multicolumn{1}{c}{\firsttone{22.59}} &
        \multicolumn{1}{c}{\firsttone{0.7161}} &
        \multicolumn{1}{c}{\secondtone{20.62}} &
        \multicolumn{1}{c}{\firsttone{0.7392}} &
        \multicolumn{1}{c}{\secondtone{21.61}} &
        \multicolumn{1}{c}{\firsttone{0.7277}} &
        \multicolumn{1}{c}{41M} \\ \bottomrule

    \end{tabular}
    }
\vspace{-0.35cm}
\end{center}
\end{table}

    \vspace{-.15in}
    \paragraph{Loss Function.} 
    We use a multi-term loss to supervise both low-level fidelity and perceptual realism during training:
    %
$
        \mathbf{L} = \alpha\mathbf{L}_{l1} + \beta\mathbf{L}_{vgg} + \gamma\mathbf{L}_{ssim} + \delta\mathbf{L}_{lpips}
$
    %

    \vspace{-.15in}
    \paragraph{Ablation Study.}
    \label{sec:ablation}
    We have performed an ablation study of our ExpoMamba model over the LOL-v1 dataset. The standard convolutional block in U-Net/M-Net architecture was replaced with \texttt{DoubleConv} block.. \texttt{Block} refers to the residual block used in each upsampling stage. We explored three HDR variants: \texttt{HDR} and \texttt{HDROut}, which use a single HDR layer placed at different locations, and \texttt{HDR-CSRNet+}, a deeper network originally designed for congested scene recognition and used within FSSB.

    The results show that \texttt{DoubleConv} and \texttt{FSSB} contribute most significantly to performance, with FSSB alone improving PSNR to 22.459. The inclusion of residual \texttt{Blocks} and the \texttt{DA} module further enhances quality. Among HDR variants, \texttt{HDR-CSRNet+} achieves the highest gains, reaching 25.640 PSNR and 0.860 SSIM, outperforming both \texttt{HDR} and \texttt{HDROut}. Each component plays a key role in maximizing the effectiveness of ExpoMamba.

\vspace{-.05in}
\section{Human Evaluation}
\label{sec:human-study}
\vspace{-.05in}

\begin{figure}[ht]
    \centering
    \includegraphics[width=0.48\textwidth]{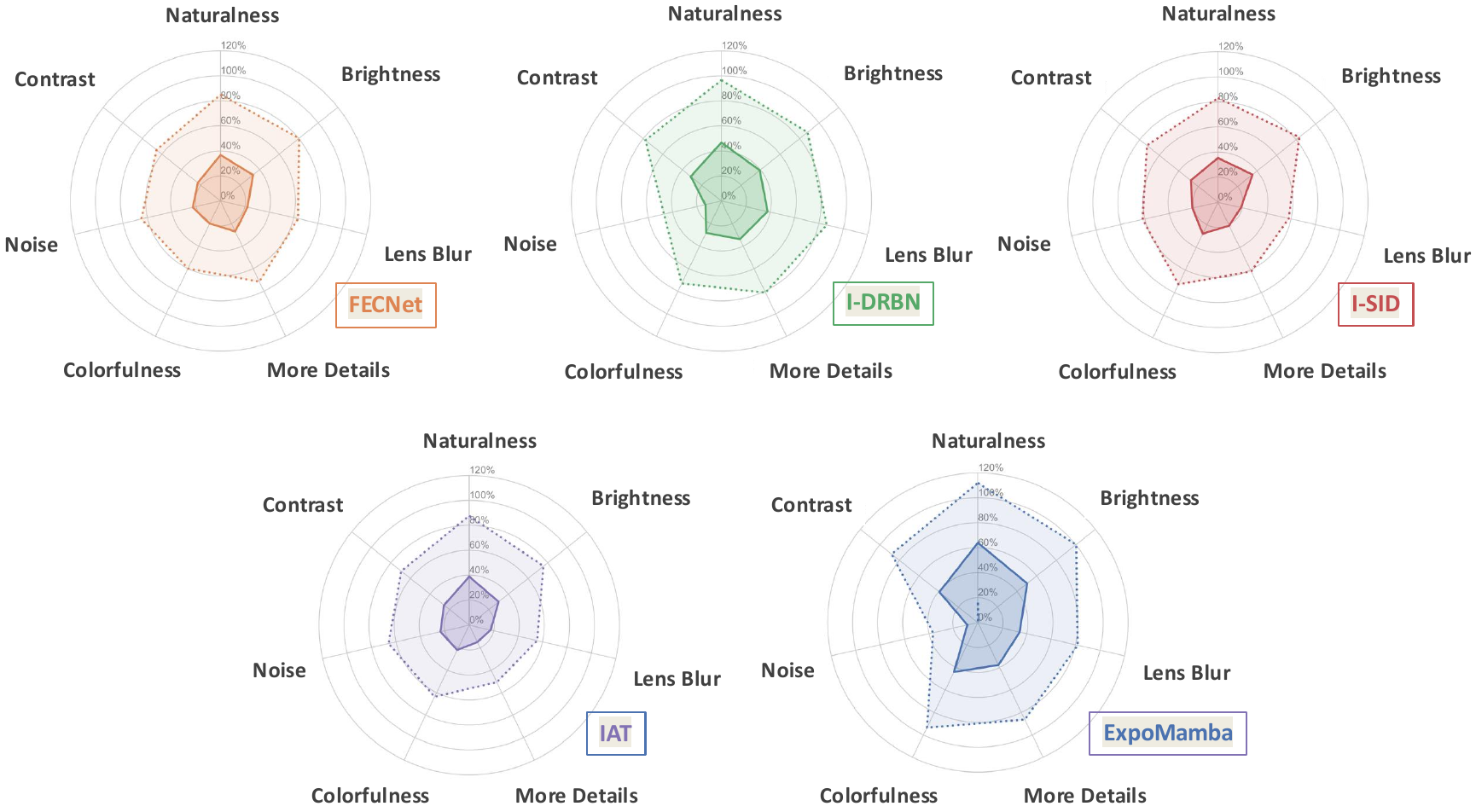}
    \caption{Perceptual evaluation results for ExpoMamba and baseline models (FECNet, I-DRBN, I-SID, and IAT) across key criteria. 
    ExpoMamba demonstrates a balanced and robust performance, particularly excelling in brightness, reduced blur, and detail retention.}
    \label{fig:overall-spider-charts}
\end{figure}

\begin{figure}
\centering
    \includegraphics[width=0.48\textwidth]{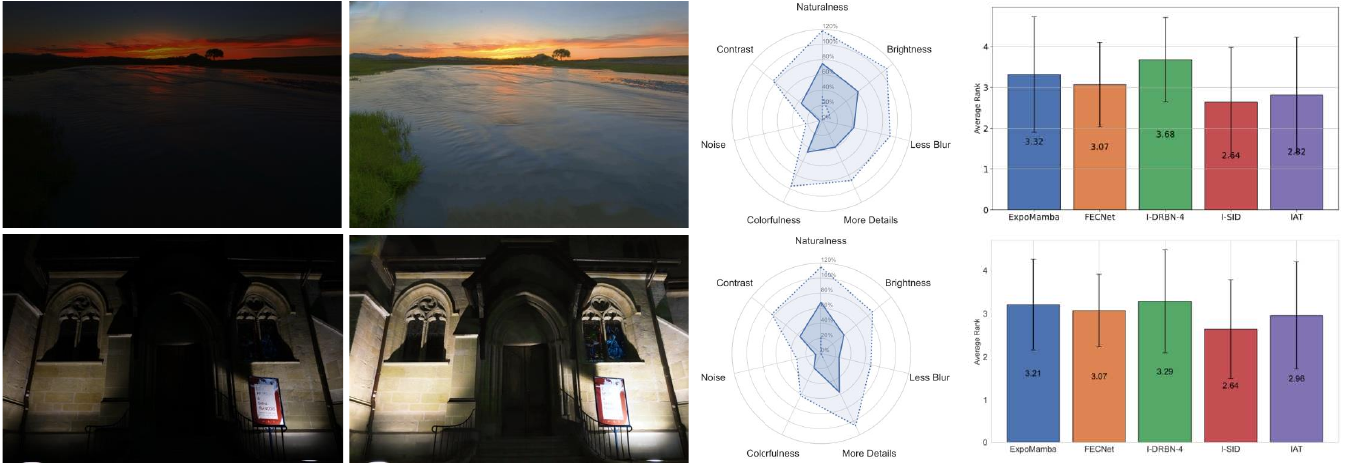}
    \caption{\label{fig:sice-test-img-1}Outputs with corresponding spider charts and average rank rating in human evaluation. The images are selected from the test dataset as the top most diverse set of images showing varying lux values, mixed-exposure, varying color density.\vspace{-.15in}}
\end{figure}

    We performed a human study with 30 participants\footnote{30 participants selected randomly from a pool of over 96,000 on the crowdsourcing platform Prolific.} and 7 perceptual criteria\footnote{We assessed~\cite{aithal2025lenvizhighresolutionlowexposurenight} each model’s naturalness, brightness, color, sharpness, detail, noise, and contrast to fully map its strengths and weaknesses.} beyond statistical metrics that may overlook perceptual aspects. Participants assessed the performance of various image enhancement models on ten diverse scenes, including varied objects, landscapes, and illuminance levels. See Figs.~\ref{fig:overall-spider-charts} and \ref{fig:sice-test-img-1}.

    From Tab.~\ref{tab:sicev2}, we picked four strong baselines: IAT \cite{Cui_2022_BMVC}, I-SID \cite{huang2022exposure}, FECNet \cite{huang2022deep}, DRBN \cite{drbn}. ExpoMamba meets or exceeds them ($3.23$ vs DRBN $3.27$, FECNet $2.92$, I-SID $2.88$, IAT $2.73$) Supp. Fig. 8, 
    yet runs 2–3$\times$ faster and uses $\sim\!4\times$ less memory. 
    These findings 
    support our ``comparable quality, better efficiency'' claim. Additional results in \textbf{Appx.\ C}.

\vspace{-.05in}
\section{Results}
\label{sec:results}
\vspace{-.025in}
    The performances in Tables~\ref{tab:lolv1-lolv2}, \ref{tab:4k}, \ref{tab:downstream-tasks}, and \ref{tab:sicev2}. 
    Tab.~\ref{tab:lolv1-lolv2} compares ExpoMamba to 19 state-of-the-art baselines, including lightweight and heavy models. We evaluate performance using \texttt{SSIM}, \texttt{PSNR}, \texttt{LPIPS}, and \texttt{FID}. ExpoMamba$_s$ achieves an inference time of \textbf{36 ms}, faster than most baselines (Fig.~\ref{fig:teaser}) and is the fastest among comparable models. Models like DiffLL \cite{jiang2023low}, CIDNet \cite{feng2024hvi}, and LLformer \cite{wang2023ultra} have comparable results but much higher inference times. Traditional algorithms (e.g., MSRCR~\cite{jobson1997multiscale}, MF~\cite{fu2016fusion}, BIMEF~\cite{ying2017bio}, SRIE~\cite{fu2016weighted}, FEA~\cite{dong2011fast}, NPE~\cite{wang2013naturalness}, and LIME~\cite{guo2016lime}) generally perform poorly on LOL4K (Tab.~\ref{tab:4k}). Fig.~\ref{fig:teaser}.b shows that increasing image resolution to 4K significantly increases inference time for transformer models due to their quadratic complexity. Despite being a 41 million parameter model, ExpoMamba demonstrates remarkable storage efficiency, consuming ${\sim}\!1/4^{th}$ memory (2923 Mb) compared to CIDNet, which, despite its smaller size of 1.9 million parameters, consumes 8249 Mb. This is because of ExpoMamba's state expansion design, which aligns computation with the GPU’s high-bandwidth memory characteristics and removes the quadratic bottleneck, thereby reducing memory footprint. 
    Current SOTA models CIDNet \cite{feng2024hvi} and LLformer \cite{wang2023ultra} are slower and less memory-efficient.

    \vspace{-.15in}
    \paragraph{Comparison with Mamba-based Methods.}
    As detailed in Tab.~\ref{tab:mamba_comparison}, ExpoMamba significantly outperforms baselines across both datasets. These results validate our core architectural hypothesis: the explicit, decoupled modeling of amplitude and phase within our FSSB provides a more effective framework for LLIE than using Mamba as a general-purpose feature extractor.

    \vspace{-.15in}
    \paragraph{Illuminating Real-World Downstream Tasks.}~To assess the real-world applicability of ExpoMamba, we evaluate its impact on downstream tasks using the ExDark dataset~\cite{Exdark} and ACDC dataset~\cite{SDV21}. Our downstream evaluations demonstrate that ExpoMamba not only produces perceptually improved images, but also enhances the operational accuracy of vision models in critical tasks of object detection and segmentation. This performance (with details shown in Tab.~\ref{tab:downstream-tasks}) suggests that {\em ExpoMamba} can serve as a robust preprocessing stage for machine-oriented pipelines, particularly in domains requiring vision reliability under poor illumination.

    \vspace{-.15in}
    \paragraph{Analysis of Efficiency.}
    A recurring question in evaluating modern architectures is the relationship between parameter count and computational speed. ExpoMamba$_s$ (41M params) has a larger parameter count than some lightweight methods, yet we report significantly faster inference times (36ms). This apparent contradiction is resolved by examining architectural efficiency rather than parameter counts.
    
    The primary speed advantage of ExpoMamba stems from the \textbf{linear-time complexity ($O(N)$)} of the underlying SSM, which fundamentally avoids the quadratic scaling bottleneck ($O(N^2)$) of self-attention in transformers. 
    This advantage becomes particularly pronounced at high resolutions. Furthermore, our model is hardware-aware, leveraging optimized \textbf{kernel fusion} techniques that significantly reduce memory I/O overhead by a factor of 20-40$\times$~\cite{gu2023mamba}, which also contributes to its lower memory footprint compared to other models. Thus, while a higher parameter count provides the necessary model capacity for quality enhancement, it is the fundamental efficiency of the SSM architecture and its hardware-conscious implementation that deliver low-latency performance.


\vspace{-.08in}
\section{Conclusion}
\label{sec:conclusion}
\vspace{-.025in}
    ExpoMamba combines frequency-based processing and state-space models to enhance low-light images, offering robust performance and efficiency for real-world applications like mobile devices and surveillance systems. Our extensive experiments, including comparisons with SOTA models and human perceptual studies, validate the superiority of our approach. 

{
    \small
    \bibliographystyle{ieeenat_fullname}
    \bibliography{main}
}

\clearpage
\newpage \newpage

\section*{Appendix}

\subsection*{A. Additional Related Work}
\label{sec:appendix_related_work}

    \textbf{Mamba Models.}~The emergence of mamba~\cite{gu2023mamba,vim} has catalyzed a new wave of innovation in low-light image enhancement (LLIE), with new architectures often building upon established theoretical frameworks. A significant portion of initial models are retinex inspired, integrating Mamba as a powerful and efficient engine for image decomposition and restoration. Pioneering works like RetinexMamba~\cite{bai2025retinexmamba} demonstrated mamba's viability by replacing computationally heavy transformer blocks to achieve greater efficiency. This was extended by LLEMamba~\cite{zhang2024llemamba}, which embeds mamba within a deep unfolding network corresponding to an ADMM optimization algorithm, making the enhancement process both interpretable and context-aware. Other models like MambaLLIE~\cite{weng2024mamballie} address the ``local pixel forgetting'' artifact of 1-D scanning by augmenting their state-space modules with local convolutions. Meanwhile, EffRetMamba~\cite{EffRetMamba} prioritizes speed by using jump-sampling to process a shorter sequence of image tokens, trading some information fidelity for a significant boost in inference time.
    
    A second category of Mamba-based models moves beyond simple backbone replacement to propose novel interaction and learning paradigms. The most radical of these is BSMamba~\cite{zhang2025bsmamba}, which discards conventional spatial scanning entirely. Instead, it reorders image patches based on brightness and semantic content before the 1-D scan, allowing the model to establish long-range dependencies between functionally related but spatially distant pixels. Other models focus on alternative enhancement strategies or learning frameworks. ResVMUNetX~\cite{wang2024resvmunetx}, for example, opts for maximum efficiency by training a Mamba network to predict a simple additive residual map, enabling real-time video enhancement. Semi-LLIE~\cite{li2024semi} tackles the practical challenge of data scarcity, using a semi-supervised framework with a mamba backbone to effectively leverage vast amounts of unlabeled data. These diverse approaches highlight the versatility of the Mamba architecture and underscore a key trend: the most successful models are not naive, drop-in replacements, but are thoughtfully designed to synergize mamba's strengths with domain-specific knowledge and innovative learning strategies.

    \textbf{Transformer Models.}~Transformer-based methods have been widely used for low-light enhancement due to their ability to model long-range dependencies. LLFormer~\cite{wang2023ultra} uses global attention to handle illumination inconsistencies. IAT~\cite{Cui_2022_BMVC} adapts lightweight transformer blocks for exposure correction. IPT~\cite{chen2021pre} employs multi-task pretraining across restoration tasks using a shared transformer backbone. Fourmer~\cite{zhou2023fourmer} integrates Fourier transforms for improved global modeling. LYT-Net~\cite{brateanu2024lyt} leverages the YUV color space for resource-efficient enhancement. MAXIM~\cite{tu2022maxim} introduces multi-axis gated MLPs for spatial and channel modeling. While effective, these models often incur high memory usage and quadratic runtime, making them less suited for real-time or mobile deployment.

    \begin{table}[t]
\begin{center}
\caption{\label{tab:model-config}Descriptions of two variants of our model, \textbf{s}' and \textbf{l}’, representing small and large model configurations.}
\resizebox{\columnwidth}{!}{
    \begin{tabular}{ccccccc}
        \toprule
        \multirow{2}{*}{\textbf{Model Type}} &
        \multicolumn{4}{c}{\textbf{Configuration}} & 
        \textbf{Inference} &
        \textbf{Memory} \\ \cmidrule(r){2-5}
        & 
        \textbf{base channel} & 
        \textbf{patch size} & 
        \textbf{depth} & 
        \textbf{params} & 
        \textbf{speed} &
        \textbf{consumption} \\ \midrule
    
        $\text{ExpoMamba}_\text{s}$ & 48 & 4 & 1 & 41 M & 36 ms & 2923 Mb\\
        $\text{ExpoMamba}_\text{l}$ & 96 & 6 & 4 & 166 M & 95.6 ms & 5690 Mb \\ \bottomrule
    \end{tabular}}
\end{center}
\end{table}

    \textbf{Diffusion Models.}~Diffusion models have shown great potential in generating realistic and detailed images. The ExposureDiffusion model \cite{wang2023exposurediffusion} integrates a diffusion process with a physics-based exposure model, enabling accurate noise modeling and enhanced performance in low-light conditions. Pyramid Diffusion \cite{zhou2023pyramid} addresses computational inefficiencies by introducing a pyramid resolution approach, speeding enhancement without sacrificing quality. \cite{saharia2022palette} handles image-to-image tasks using conditional diffusion processes. Models like \cite{zhang2022gddim} and deep non-equilibrium approaches \cite{pokle2022deep} aim to reduce sampling steps for faster inference. However, starting from pure noise in conditional image restoration tasks remains a challenge for maintaining image quality while cutting down inference time \cite{guo2023shadowdiffusion}.

    \textbf{Hybrid Modeling.}~Hybrid models include learning features in both spatial and frequency domains have been another popular area in image enhancement/restoration tasks. It has been predominantly explored in three sub-categories: \textbf{(1)}~ Fourier Transform \cite{yuan2024multi}, Fourmer \cite{zhou2023fourmer}, FD-VisionMamba \cite{zheng2024fdvision}; \textbf{(2)}~ Wavelet Transform \cite{zou2024wave, tan2024wavelet, chen2024mwrd}; and, \textbf{(3)}~ Homomorphic Filtering \cite{al2022low}. Such methods demonstrate that leveraging both spatial and frequency information can significantly improve enhancement performance. Recent hybrid models such as MAXIM \cite{tu2022maxim} and PromptIR \cite{potlapalli2023promptir} have explored lightweight but flexible design spaces for image restoration and enhancement tasks.

\subsection*{B. Extended Details on Phase Manipulation}
\label{ex-phase-man}

    This section provides a step-by-step derivation of how a uniform phase shift in the Fourier domain affects an image in the spatial domain.
    
    We begin with the definition of the 2D Fourier Transform for an image $I(x, y)$:
    \begin{equation}
        \mathbf{F}(u, v) = \mathcal{F}\{I(x, y)\} = \iint_{-\infty}^{\infty} I(x, y) e^{-i2\pi(ux + vy)} \,dx\,dy
    \end{equation}
    The transform can be represented in polar form using its amplitude $\mathbf{A}(u, v)$ and phase $\phi(u, v)$:
    \begin{equation}
        \mathbf{F}(u, v) = \mathbf{A}(u, v) e^{i\phi(u, v)}
    \end{equation}
    The original image $I(x, y)$ is recovered via the Inverse Fourier Transform:
    \begin{equation}
        I(x, y) = \mathcal{F}^{-1}\{\mathbf{F}(u, v)\} = \iint_{-\infty}^{\infty} \mathbf{F}(u, v) e^{i2\pi(ux + vy)} \,du\,dv
    \end{equation}
    Substituting the polar form into the inverse transform gives:
    \begin{equation}
        I(x, y) = \iint_{-\infty}^{\infty} \mathbf{A}(u, v) e^{i\phi(u, v)} e^{i2\pi(ux + vy)} \,du\,dv
    \end{equation}
    Our goal is to analyze the effect of applying a uniform phase shift, $\Delta\phi$, to the phase component, resulting in a new phase $\phi'(u, v) = \phi(u, v) + \Delta\phi$. The modified Fourier spectrum is $\mathbf{F'}(u, v) = \mathbf{A}(u, v) e^{i(\phi(u, v) + \Delta\phi)}$. The corresponding image in the spatial domain, $I'(x, y)$, is derived as follows.
    
    \begin{align}
        I'(x, y) &= \iint \mathbf{A}(u, v) e^{i(\phi(u, v) + \Delta\phi)} e^{i2\pi(ux + vy)} \,du\,dv \label{eq:start} \\
        &= \iint \mathbf{A}(u, v) e^{i\phi(u, v)} e^{i\Delta\phi} e^{i2\pi(ux + vy)} \,du\,dv \label{eq:separate_exp} \\
        &= e^{i\Delta\phi} \iint \mathbf{A}(u, v) e^{i\phi(u, v)} e^{i2\pi(ux + vy)} \,du\,dv \label{eq:factor_out} \\
        &= e^{i\Delta\phi} \cdot I(x, y) \label{eq:substitute_I} \\
        &= (\cos(\Delta\phi) + i\sin(\Delta\phi)) \cdot I(x, y) \label{eq:euler} \\
        &= \cos(\Delta\phi) I(x, y) + i\sin(\Delta\phi) I(x, y) \label{eq:final}
    \end{align}
    
    \vspace{0.5em}
    \noindent In this derivation: 
    Eq.~\eqref{eq:separate_exp} uses the exponential identity $e^{a+b} = e^a e^b$; 
    Eq.~\eqref{eq:factor_out} factors out the constant phase shift $\Delta\phi$ from the integral; 
    Eq.~\eqref{eq:substitute_I} recognizes the inverse Fourier transform of the original image; 
    Eq.~\eqref{eq:euler} applies Euler’s formula $e^{i\theta} = \cos\theta + i\sin\theta$. 
    Finally, Eq.~\eqref{eq:final} shows that a uniform phase shift in the frequency domain results in a complex-valued image where the real and imaginary parts are scaled versions of the original.
    This final expression \eqref{eq:final} demonstrates that a uniform phase shift in the frequency domain results in multiplying the original image $I(x, y)$ by a complex constant $e^{i\Delta\phi}$. The new image $I'(x, y)$ is a complex-valued signal where the real part is the original image scaled by $\cos(\Delta\phi)$, and the imaginary part is the original image scaled by $\sin(\Delta\phi)$. This rotation in the complex plane fundamentally alters the image's spatial characteristics, underscoring the critical role of phase information in defining structural content.

\subsection*{C. Additional Human Evaluation Results}
    \begin{figure}[h]
        \centering
        \caption{\label{fig:sice-test-img-others}Additional results from the SICE test set for human evaluation (continued from Fig.~\ref{fig:sice-test-img-1}), along with corresponding spider charts and average rank ratings.}
        \includegraphics[width=0.485\textwidth]{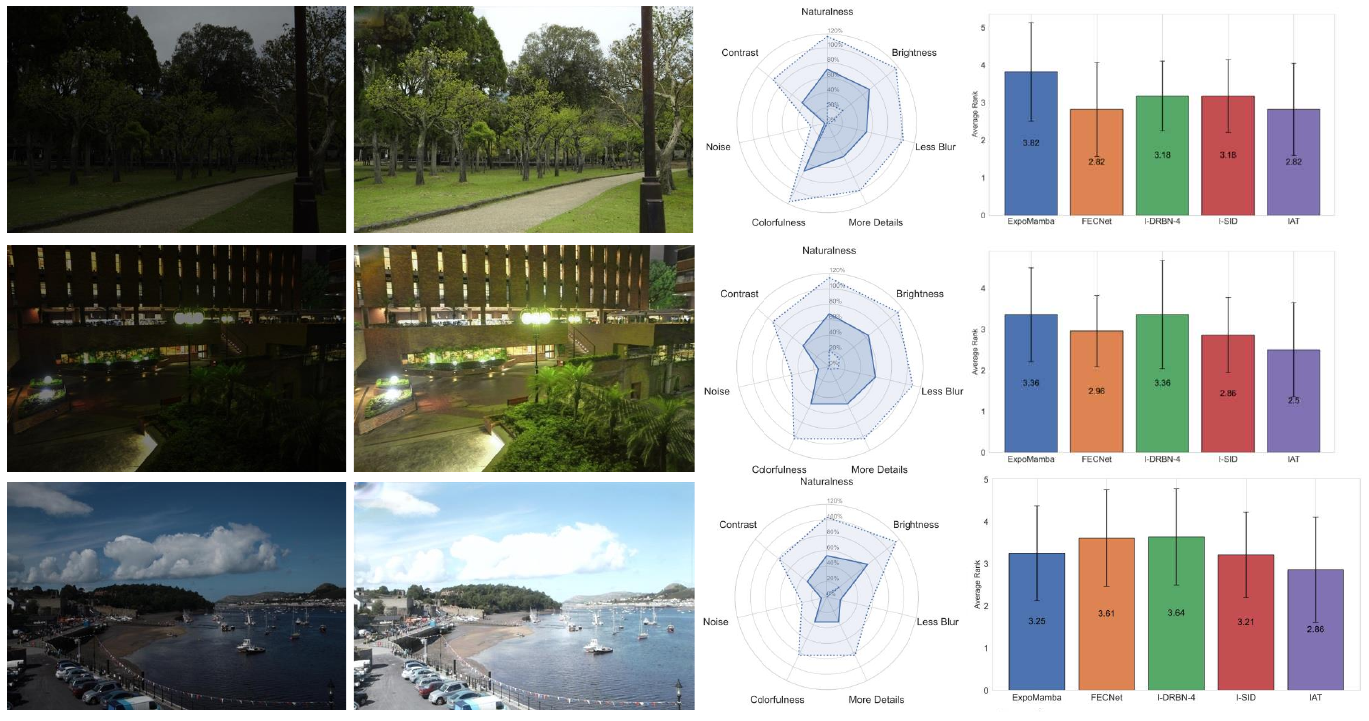}
        \includegraphics[width=0.495\textwidth]{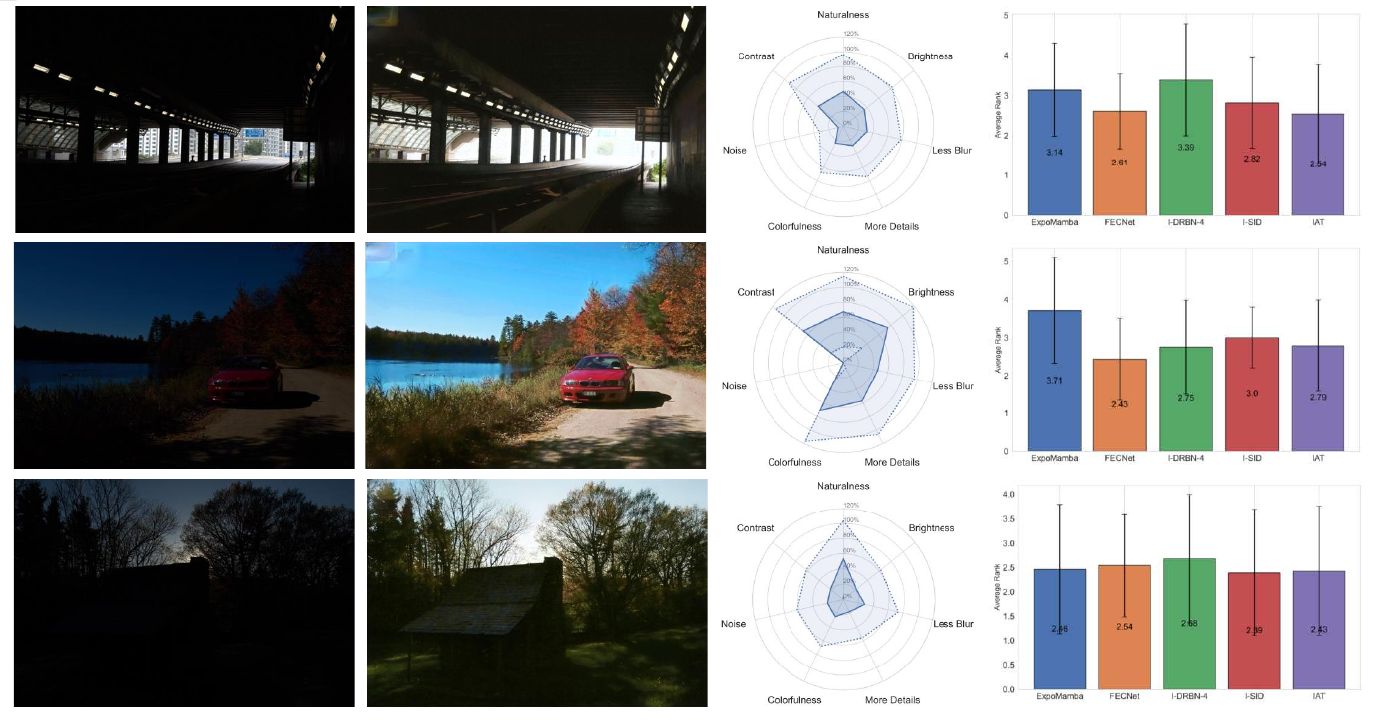}
        \includegraphics[width=0.485\textwidth]{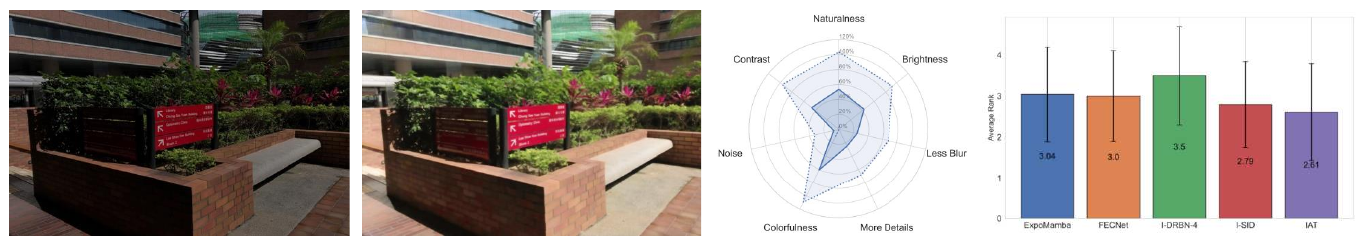}
    \end{figure}
    
    \begin{figure}[ht]
        \centering
        \caption{\label{fig:overall-rank}Aggregate rank comparison of ExpoMamba and baselines from human perceptual evaluation. Higher values indicate better subjective quality; error bars represent standard deviation.}
        \includegraphics[width=0.49\textwidth]{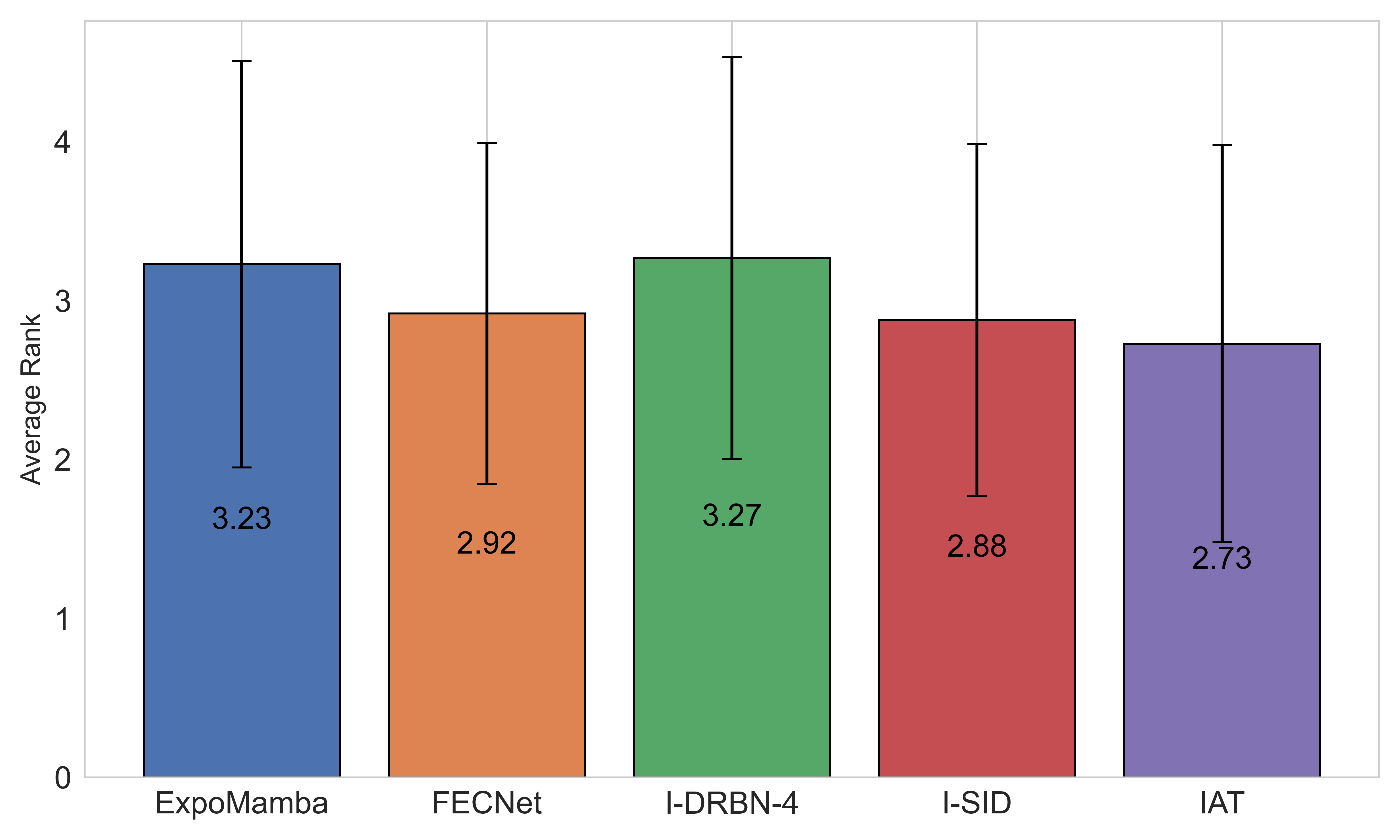}
    \end{figure}

    As an extension to the main paper’s perceptual study, we present qualitative results and statistical rankings from our human evaluation conducted on the SICE test dataset. Participants rated multiple image enhancement models across seven visual attributes: \textit{naturalness}, \textit{brightness}, \textit{colorfulness}, \textit{reduced blur}, \textit{more details}, \textit{noise reduction}, and \textit{contrast}. In the spider charts (Fig.~\ref{fig:sice-test-img-others}), ExpoMamba consistently outperforms competing methods across all criteria, with especially strong scores in the \textit{noise reduction}, \textit{naturalness}, and \textit{detail retention} dimensions.
    
    The aggregated average ranks across all images and participants are summarized in Fig.~\ref{fig:overall-rank}. ExpoMamba receives the highest overall perceptual ranking, demonstrating its ability to deliver visually balanced and subjectively preferred results. The small standard deviation further reflects strong inter-rater consistency, reinforcing the perceptual robustness of the proposed method.


    

\subsection*{D. Frequency-Dependent Dynamic Adaptation.}
    
    Let $r(u,v)=\frac{\sqrt{u^{2}+v^{2}}}{r_{\max}}$ denote the normalized radial frequency. We parameterize frequency dependent state matrices via shallow gating functions $\alpha_{A}(r)$, $\alpha_{B}(r)$, $\alpha_{C}(r)$ produced by a 1\,layer MLP with Softplus, shared across channels:
    \[
        \alpha_{\bullet}(r)
        = \log\!\Big(1 + \exp\Big(
        \mathbf{w}_{\bullet 2}\,
        \operatorname{ReLU}\!\big(\mathbf{w}_{\bullet 1}\, r + b_{\bullet 1}\big)
        + b_{\bullet 2}
        \Big)\Big).
    \]
    {where, $\bullet\in\{A,B,C\}$. For each step $t$ and frequency bin $(u,v)$, we form
    \[\begin{aligned}
        A_{t}(u,v) &= A_{t}\,\circ\,\big(1 + \alpha_{A}(r(u,v))\big),\\
        B_{t}(u,v) &= B_{t}\,\circ\,\big(1 + \alpha_{B}(r(u,v))\big),\\
        C_{t}(u,v) &= C_{t}\,\circ\,\big(1 + \alpha_{C}(r(u,v))\big).
    \end{aligned}\]
    Here, $\circ$ denotes element wise multiplication. This keeps the SSM core intact while allowing smooth low to high frequency reweighting with three learnable gates.
    After separate processing through state-space models, the modified amplitude $\mathbf{A''(u,v)}$ and phase $\mathbf{P''(u, v)}$ are recombined and transformed back into the spatial domain to reconstruct the enhanced image: $\mathbf{I'(x,y)} = \mathbf{F^{-1}} (\mathbf{A''(u.v)} + i\cdot \mathbf{P''(u.v)})$; where, $\mathbf{F^{-1}}$ denotes the inverse Fourier Transform.

\vspace{0.5em}
\subsection*{E. Dynamic Patch Training} \label{fig:dynamic_patch_training}
    \begin{figure}[b]
        \centering
        \caption{I\label{fig:dynamic-path-training}llustration of dynamic patch training: multiple resolutions are randomly batched with padding, enabling the model to generalize across scales.}
        \includegraphics[width=0.49\textwidth]{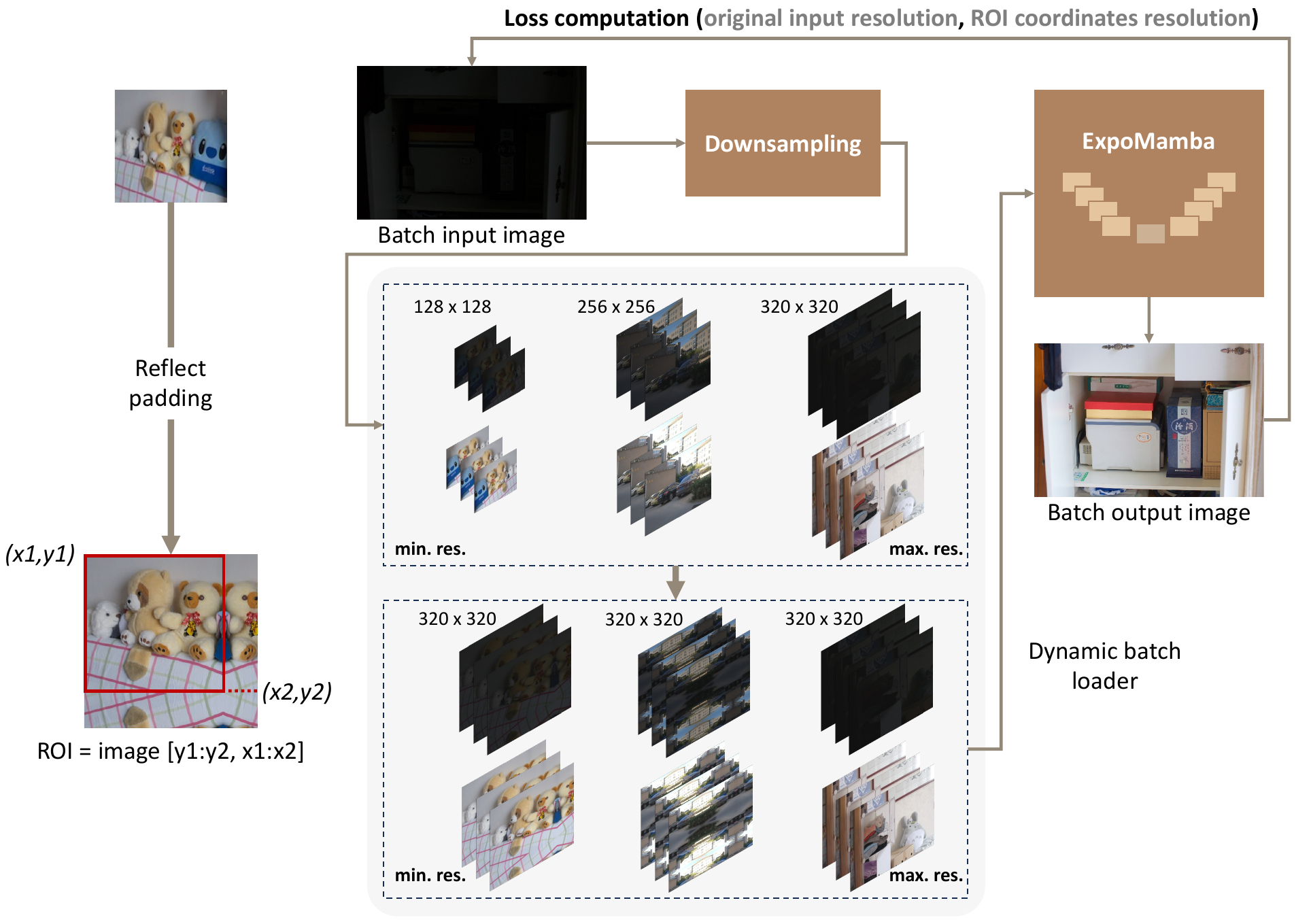}
    \end{figure}
    
    To improve robustness to variable input resolutions, ExpoMamba adopts a dynamic patch training strategy, as shown in Fig.~\ref{fig:dynamic-path-training}. Specifically:
    
    \begin{itemize}
        \item In each training batch, a random patch size is selected from $\{128^2, 256^2, 324^2\}$.
        \item Each mini-batch contains images of the same resolution, but resolution varies across batches.
        \item The 2D scanning mechanism within FSSB is thereby trained to handle multi-scale representations efficiently.
    \end{itemize}
    
    This improves generalization to real-world conditions, especially on mobile devices and webcams that adapt resolution dynamically to conserve power or bandwidth.

\subsection*{F. Dynamic Adjustment Approximation}
\label{da-approx}

    The Dynamic Adjustment Approximation (DAA) module provides an unsupervised mechanism for exposure correction by leveraging intrinsic image statistics, eliminating the need for reference ground truth maps or pre-computed illumination priors. Unlike prior approaches such as KinD, LLFlow, or RetinexFormer, which rely heavily on ground-truth mean score as guidance signals derived from paired datasets, our method is entirely self-reliant and dynamically adapts to the brightness characteristics of each input. This approach helps in boosting the performance during inference.
    
    Given an input image $\mathbf{I} \in \mathbb{R}^{C \times H \times W}$, we compute two summary statistics across all pixels and channels: the mean $\mu$ and the median $m$. A normalized intensity value $\tau \in [0, 1]$ is selected as the desired luminance anchor (empirically $\tau = 0.345$ in our case). The goal is to shift the median pixel values toward this anchor in a manner weighted by their deviation from the image's current mean brightness.
    
    The adjustment factor $\mathbf{F}$ is computed as:
    \begin{equation}
        \mathbf{F} = \frac{m + \alpha \cdot (\tau - \mu)}{m}
    \end{equation}
    
    Here, $\alpha$ is a tunable strength parameter that controls the degree of adjustment applied per image. The final adjusted image $\hat{\mathbf{I}}$ is computed element-wise as:
    \begin{equation}
        \hat{\mathbf{I}} = \mathbf{I} \times \mathbf{F}
    \end{equation}
    
    This formulation ensures stability even under extreme low-light conditions by avoiding division by near-zero values and maintains exposure balance without introducing unnatural contrast. Because the adjustment is performed in a single-pass, the method is highly efficient and can be seamlessly integrated into real-time inference pipelines. Its effectiveness is especially pronounced in deployment scenarios where reference illumination statistics are unavailable, such as mobile or embedded imaging systems.

\subsection*{G. Additional Benchmark Visualizations}
    \begin{figure*}[!ht]
    \centering
    \caption{\label{fig:lolv1}Qualitative comparison of ExpoMamba and baselines on the LOLv1 dataset. Results demonstrate structural fidelity and color balance under mixed lighting conditions.}
    \includegraphics[width=0.9\linewidth]{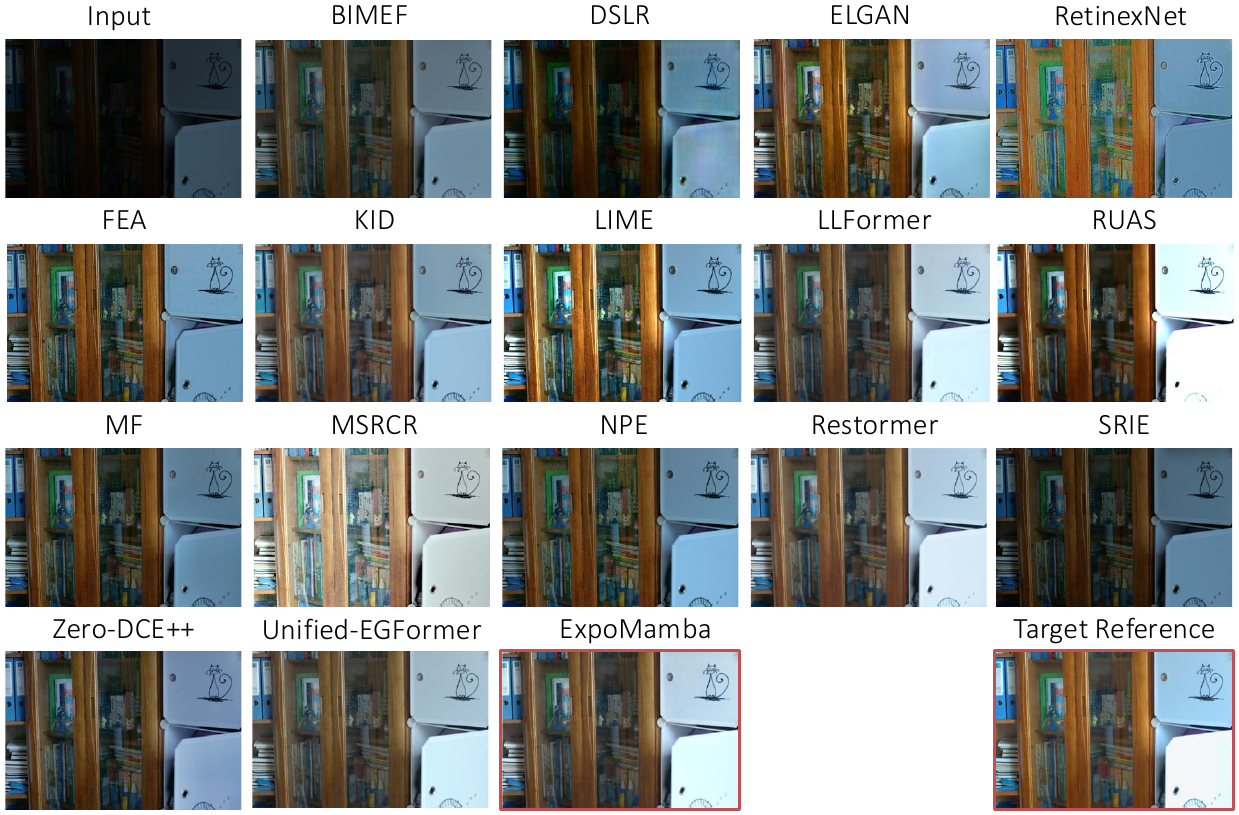}
    \end{figure*}

    Fig.~\ref{fig:lolv1} illustrates additional visual comparisons across 16 baseline methods and the proposed ExpoMamba model. While methods such as Zero-DCE++, RUAS, and LIME tend to either over-enhance or introduce unnatural hues, ExpoMamba preserves both local structure and global illumination. Compared to LLFormer and Restormer, which occasionally oversmooth textures or lose highlight fidelity, ExpoMamba maintains sharper edges and natural luminance transitions. These results further validate the model's ability to generalize well across diverse lighting conditions present in the LOLv1 dataset.

\subsection*{H. Discussion: Inference Time vs. FLOPs}

    In this work, we prioritize reporting real inference time over theoretical FLOPs, as the former provides a more accurate reflection of practical deployment efficiency. While FLOPs serve as a platform-agnostic metric for algorithmic complexity, they fail to capture critical system-level considerations such as memory bandwidth, caching behavior, and inter-layer communication, all of which play a substantial role in runtime performance on real hardware. In contrast, actual inference time directly reflects the influence of design choices such as data flow optimization, activation reuse, and parallel execution scheduling.
    
    This distinction becomes especially important in latency-sensitive applications like augmented reality, autonomous navigation, or mobile imaging, where wall-clock latency, not theoretical compute, determines responsiveness and usability. For ExpoMamba, we report inference time measured on an NVIDIA A10G GPU, which accounts for the full end-to-end processing pipeline, including frequency decomposition, dual VSSM inference, and reconstruction. This emphasis on timing enables a fairer and more relevant comparison with baseline models in the context of deployment scenarios.
    
    Optimizing for real-time execution rather than abstract operation counts ensures that ExpoMamba is not only computationally efficient but also pragmatically viable for edge environments where latency, memory, and energy constraints coexist.

\subsection*{I. Extended Comparative Efficiency and Model Scalability}

    \textbf{Comparative Efficiency Analysis.} ExpoMamba achieves a strong balance between visual quality and practical deployment metrics. As reported in Tab.~\ref{tab:lolv1-lolv2}, it processes a $400 \times 600$ image in 36 ms with a 2923 MB memory footprint—substantially faster than DiffLL~\cite{jiang2023low} (158 ms, 8249 MB) and LLFormer~\cite{wang2023ultra} (1956 ms, >6 GB). This performance highlights the advantage of using linear-time operations and frequency-state modeling over transformer-based alternatives for real-time applications.
    
    \textbf{Balancing Speed and Effectiveness.} Although ExpoMamba is not the smallest model in terms of parameter count, it outperforms many smaller baselines (e.g., IAT: 0.09M, FECNet+ERL: 0.15M) in both perceptual enhancement and downstream task accuracy (Tab.~\ref{tab:downstream-tasks}). Its modular architecture supports scalable deployment: ExpoMamba\textsubscript{s} offers a lightweight configuration for edge devices, while ExpoMamba\textsubscript{l} can scale up for cloud or desktop inference. This adaptability—combined with low latency and generalization across tasks—makes it a strong candidate for real-world LLIE pipelines in mobile, surveillance, and automotive domains.

\subsection*{J. Expanded Algorithm for ExpoMamba}
    \begin{algorithm}[!ht]
    \small
    \caption{\label{alg:full}ExpoMamba Training with Frequency State Space Block (FSSB)}
    \begin{algorithmic}[1]
    \STATE \textbf{Input:} Dataset $\mathcal{D}$, Training epochs $E$, Components: FSSB, VSSM\textsubscript{A}, VSSM\textsubscript{P}, HDR, ComplexConv
    \STATE \textbf{Output:} Optimized model parameters $\theta$
    \\
    \STATE \textbf{// Frequency Decomposition}
    \FOR{\textbf{each image} $I \in \mathcal{D}$}
        \STATE Compute Fourier Transform $\mathcal{F}(u,v) = \mathcal{F}[I(x,y)]$
        \STATE Decompose: $\mathcal{F}(u,v) \rightarrow A(u,v), P(u,v)$
    \ENDFOR
    \\
    \STATE \textbf{// Frequency-State Modeling (FSSB)}
    \FOR{\textbf{each component} $(u,v)$}
        \STATE Update VSSMs: \\$\mathbf{h}[t+1] = \mathbf{A}[t]\cdot\mathbf{h}[t] + \mathbf{B}[t]\cdot\mathbf{x}[t]$, \quad $\mathbf{y}[t] = \mathbf{C}[t]\cdot\mathbf{h}[t]$
        \STATE Generate modulated outputs $A''(u,v), P''(u,v)$
    \ENDFOR
    \\
    \STATE \textbf{// Inverse Transform \& Reconstruction}
    \STATE Combine frequency outputs: $\hat{\mathcal{F}}(u,v) = A''(u,v) + i \cdot P''(u,v)$
    \STATE Apply Inverse Fourier Transform: $\hat{I}(x,y) = \mathcal{F}^{-1}[\hat{\mathcal{F}}(u,v)]$
    \\
    \STATE \textbf{// Model Training}
    \FOR{$e = 1$ to $E$}
        \FOR{\textbf{each batch} $\mathcal{B} \subset \mathcal{D}$}
            \STATE Pass $\mathcal{B}$ through FSSB $\rightarrow$ HDR $\rightarrow$ ComplexConv
            \STATE Compute loss $\mathcal{L}$; Backpropagate $\nabla_{\theta} \mathcal{L}$
        \ENDFOR
    \ENDFOR
    \STATE \textbf{Return:} Trained parameters $\theta$
    \end{algorithmic}
    \end{algorithm}

\begin{figure*}[!ht]
    \centering
    \includegraphics[width=\textwidth]{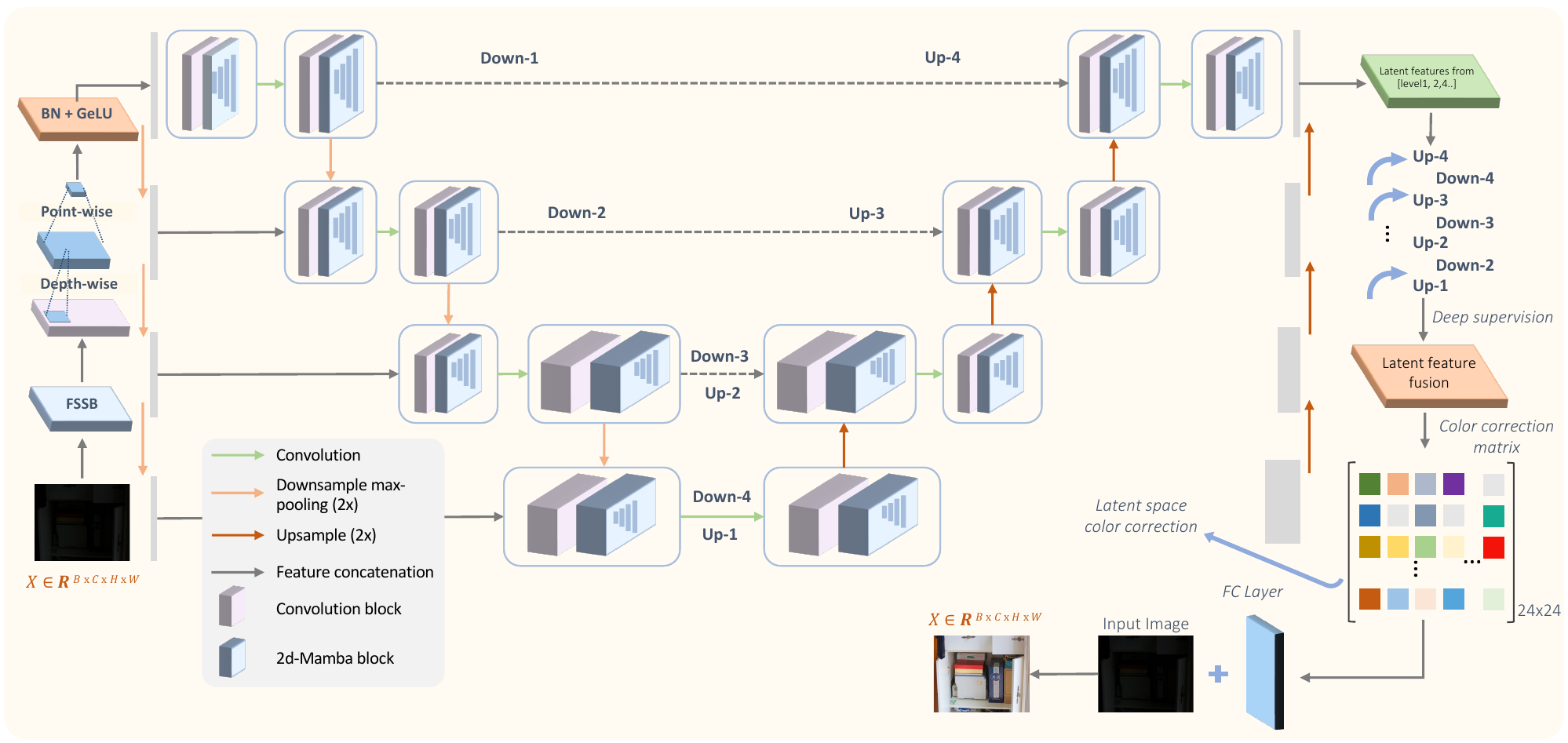}
    \caption{\label{fig:expomamba-arch-supp} Overview of the ExpoMamba Architecture. The diagram illustrates the information flow through the \emph{ExpoMamba} model.}
\end{figure*}

\subsection*{K. Model Configuration.}
    Model configuration (see Table \ref{tab:model-config} for details) provides a detailed comparison between the two variants of  ExpoMamba, highlighting their configurations and performance metrics. Notably, despite an increase of 125 million parameters, the memory of the larger $\text{ExpoMamba}_{\text{l}}$ variant is 5690 Mb, which is a modest increase compared to transformer-based models and $\text{ExpoMamba}_{\text{s}}$.

\end{document}